# Algebraic Net Class Rewriting Systems, Syntax and Semantics for Knowledge Representation and Automated Problem Solving

**Seppo Ilari Tirri**

*PhD (ICT) student, Information & Communication Technology*
*Asia e University (AeU), Malaysia*
*Jalan Sultan Sulaiman, 50000 Kuala Lumpur, Malaysia*

`seppo.tirri@aeu.edu.my`

ABSTRACT

The intention of the present study is to establish general framework for automated problem solving by approaching the task universal algebraically introducing knowledge as realizations of generalized free algebra based nets, graphs with gluing forms connecting in- and out-edges to nodes. Nets are caused to undergo transformations in conceptual level by type wise differentiated intervening net rewriting systems dispersing problems to abstract parts, matching being determined by substitution relations. Achieved sets of conceptual nets constitute congruent classes. New results are obtained within construction of problem solving systems where solution algorithms are derived parallel with other candidates applied to the same net classes. By applying parallel transducer paths consisting of net rewriting systems to net classes congruent quotient algebras are established and the manifested class rewriting comprises all solution candidates whenever produced nets are in anticipated languages liable to acceptance of net automata.

# INTRODUCTION

## MOTIVATION

In all fields of data processing, especially in robotics, physics and overall changing constructions is ever increasing need for knowledge of common structures in creating fast, exact, controllable and sufficiently comprehensive solving algorithms for problems. From René Descartes freely quoted: "there is not very much in results or even in the proofs of them, but the method how they are invented, that is what is the process inventors use to realize proofs". Restricting data flow to finite cases is often in descriptive models improper in order to get sufficient model to handle with the tasks, e.g. if variables are allowed to be systems themselves as in function representatives of quantum particles. Models in meteorology and models for handling with populations, biological organizations or even combinations in genetic codes call for common approach in problem solving especially in cases where in- or out- data flow volumes are beforehand impossible to predict to be limited in the already known sphere. For connections between neurons in brains, and in more theoretical aspects for allowance of simultaneous "loops", nets are ideal as formal representations for iterations as e.g. within solutions for powers of higher order differential equations by Picard successive iterants. In robotics strong AI, the abstract mathematical reasoning model, will play the key element in handling data processes in artefacts. In the 1980s in Japan were the first concrete steps taken in robotics trying to imitate human actions; however imitating process is the endless effort to achieve inventiveness which lays its solid ground in the understanding of reasoning itself, thus strong AI is a more effective approach. Infinite ranks (the mightiness or cardinality of in- or out places in operation relations) are needed as tools for infinite simultaneous data flow into systems (operations) such as in quantum physics where infinite number of different state function solutions of a Schrödinger-equation compounds a field to be operated. Naturally one can imagine numerous other fields where a mathematical framework for problem solving would be desirable. Within solving any problem an essential thing is to see over details, and one inevitably confronts the necessity of outlining or abstracting the object to be solved to already more familiar forms or to forms easier to be checked – keeping the number of links regarding the environment of the object in hands unchanged (e.g. to be the most comprehensive, the handled data flow would not be allowed to be restricted solely to beforehand computably predictable form).

USED TOOLS

The notion of net is introduced as the formal representative for comprehensive idea of information. In order to define comprehensive, mathematically sustaining definition for the conceptual problems we use pairs comprising problem objects and recognizers, where those object concerned are nets, namely graphs (allowing loop structures) where vertices (nodes) are ranked letters glued by arity letters to each other at their arity places in the procedure: in to out and out to in. Recognizers can choose to be any netmorphism or even rewriting systems or transducers (graph-like formations the nodes replaced by rewriting systems).

To avoid beforehand unnecessarily restricting unpredictable data to include in problem objects we choose to use infinite formulation in object definition, that means allowing unlimited number of in- and output places to be occupied in nets. Hence the used exposure for objects comprises models for phenomena crucially vulnerable to unpredictable data flow.

Therefore we introduce three types of alphabets (sets of letters or signature) distinct from each other: frontier alphabet (the letters called variables), ranked alphabet (the letters called operators) and arity alphabet (of arity letters) divided in in-arity and out-arity alphabets, distinct from each other. The distinction for in- and out-arities is needed to divide the data flow direction in operations.

Realization (valuation) of nodes shifts the ranked letters (operators) to operations (relations: deterministic or undeterministic (with many valued) functions) which are operational in a desired algebra considering the practice at hand. In the special case of terms the operators can be regarded as free algebra operations. Realizations are defined so that the coming (in upstream) node operators are allowed to influence to the images of the node operations concerned. That is essential in loop structures.

Equations and decompositions are presented in the most general course for the purpose of closure properties in the realm of net classes, following systematically chosen way to dress information to suit to nets and accordingly to transducers.

METHOD AND TARGET

Various operations among nets are presented as well as rewriting systems, essential to comprehend used derivations. E.g. solving equation groups by the replacing method falls into partition structures.

The notion of inventiveness is probed: Comprehensive general exact solving method and its characteristic features especially in quotient systems obtained by partition or more generally by cover rewriting systems (a generalization for the idea of partition, consisting of the depth dimension of conceptual rewriting liable to allow left and right sides of rules intersect with each other) are targeted. We use parallel rewriting, commuting rectangles and iterations and in the cases of prerequisites related to limit demands, check results by quotient automata where the final states are sets of class elements. Systems have resemblances with confluence properties within classes and natural transformation between Functors demonstrating parallel algebras.

Rewriting systems will be classified type wise, comprising also the possibility to use the left and on the other hand distinctively the right substitutions. By right side substitution relations new links can be created from the applicable object to unoccupied in- or output places in the particular object itself. Hence we can in some cases avoid using infinite number of rewrite rules. Class characterizations are given to ease the burden of formatting abstract pairs.

OVERALL SUMMATION OF WORKING ORDER

The present work is offering an explicit model for the representation of knowledge itself, not only implicit often seen in graph definitions and graph rewriting. Furthermore we deliver explicit syntax for automated problem solving.

The work is basing itself on a generalization and transformation to universal algebraic net configuration with new realization definition and renetting types on. First we present necessary preliminary definitions for the construction of nets the nodes of which have arbitrary number of in- and outputs. Realizations of nets are defined by transformations from operators to operations in algebras. Then we give the type wise representation for renetting systems and transducers, the node realizations of which being rewriting systems. The necessary consideration is given to definitions for generalized equations for closure properties in net class rewriting. The definition of problem and its solution is introduced in terms of nets, recognisability and transducers fulfilling limit demands. Then the partition of nets and the abstraction relation between concept nets are introduced yielding net classes, needed in searching the fitting partial solutions from memory.

"Altering macro renetting system"-theorem is introducing the necessary equation matching each step of the solution process between the substances of the nets in jungles. Parallel theorem establishes the invariability of the abstraction relation and also the construction for necessary algorithms for abstract sisters subject to net class rewriting algebra. The construction process of the desired transducer for the jungles in given problems to be solved is thus obtained from the known ones in iteratively updated memory. Finally we present the extension of the rules of solving transducers, in the cases where covers of mother nets in problems differ from partitions, where cover renetting systems are defined as generalizations of partition ones, and notion partition of jungle is replaced by concept of cover renetting result consisting of sequential parts of cover in depth dimension, partly replaced by each other.

# 1.§ Preliminaries

This work follows the general custom of the discipline in concern and only neccessary symbol definitions are manifested, readers are encouraged to turn to the literature represented in the reference list for the more comprehensive guidance.

## 1.1. Sets and Relations

We agree that all defined terms are of the cursive style when represented first time.

**Definition 1.1.01.** We regularly use small letters for elements and capital letters for sets and when necessary bolded capital letters for families of sets. The new defined terms are underlined when represented the first time.

**Definition 1.1.02.** We use the following convenient symbols for arbitrary element a and set A in the meaning:

$a \in A$ " a is an element of A or belongs to A or is in A "

$a \notin A$ " a does not belong to A "

$\exists \, a \in A$ " there is such an element a in A that "

$\exists! \, a \in A$ " there is exactly one element a in A "

$\nexists \, a \in A$ " there exists none element a in A "

$\forall a \in A$ " for each a belonging to A "

$\Rightarrow$ " then it follows that "

$\Leftrightarrow$ " if and only if " , shortly " iff "

**Definition 1.1.03.** {a : *} or (a : *) means a *conditional* set, the set of all such a-elements which fulfil each condition in sample * of conditions, and *nonconditional*, if sample * does not contain any condition concerning a-elements.

**Definition 1.1.04.** $\varnothing$ means *empty set*, the set with no elements. A set of sets is called a *family*. For set $\mathcal{I}$ the notation $\{a_i : i \in \mathcal{I}\}$ is an *indexed set* (over $\mathcal{I}$). Set $\{a_i : i \in \mathcal{I}\}$ is $\{a\}$, if $a_i = a$ whenever $i \in \mathcal{I}$. If there is no danger of confusion we identify a set of one element, *singleton*, with its element. It is noticeable that $\{\varnothing\}$ is a singleton set.

**Definition 1.1.05.** For arbitrary sets A and B we use the notations:

$A \subseteq B$ or $B \supseteq A$ " A is a *subset* of B (is a part of B or each element of A is in B) or B *includes* A "

$A \nsubseteq B$ " A is not a part of B (or there is an element in A which is not in B)"

$A \subset B$ or $B \supset A$ " A is a *genuine subset* of B " meaning " $A \subseteq B$  and  $(\exists\, b \in B)\; b \notin A$ "

$A \not\subset B$ " A is not a genuine subset of B "

$A \neq B$ " A is not the same as B "

$A^c$ or $\neg A$ " is the *complement* of A " meaning set $\{a : a \notin A\}$

$A \cup B$ " the *union* of A and B " meaning set $\{a : a \in A \text{ or } a \in B\}$

$A \cap B$ " the *intersection* of A and B " meaning set $\{a : a \in A, a \in B\}$. If $A \cap B = \varnothing$, we say that A and B are *distinct* with each other, or *outside each other*.

$A \setminus B$ " A excluding B " meaning $\{a : a \in A, a \notin B\}$.

**Definition 1.1.06.** The *cardinality* of A, "the number" of the elements in set A, is denoted by |A|.

**Definition 1.1.07.** P(A) symbolizes the family of all subsets of set A.

**Definition 1.1.08.** The set of natural numbers $\{1,2,...\}$ is denoted by symbol $\mathbb{N}$, and $\mathbb{N}_0 = \mathbb{N} \cup \{0\}$. Maximum of the numbers in subset A of $\mathbb{N}_0$ is denoted maxA.

**Definition 1.1.9.** Notice that for sets $A_1$ and $A_2$ and samples of conditions $*_1$ and $*_2$

$$\{a : a \in A_1, *_1\} \subseteq \{a : a \in A_2, *_2\},$$

if $(A_1 \subseteq A_2$ and $*_1 = *_2)$ or $(A_1 = A_2$ and $*_2 \subseteq *_1)$.

**Definition 1.1.10.** The notation $\cup(A_i : i \in \mathcal{I})$ is the *union* $\{a : (\exists i \in \mathcal{I})\ a \in A_i\}$ and

$$\cap(A_i : i \in \mathcal{I}) \text{ is the } \textit{intersection } \{a : (\forall i \in \mathcal{I})\ a \in A_i\}.$$

for indexed family $\{A_i : i \in \mathcal{I}\}$. For any family $\mathcal{B}$ we define:

$$\cup \mathcal{B} = \cup(\ B : B \in \mathcal{B}\ )$$

$$\cap \mathcal{B} = \cap(\ B : B \in \mathcal{B}\ ).$$

**Definition 1.1.11.** Set $\rho$ of ordered pairs (a,b) is a *binary relation* (shortly relation), where a is a $\rho$-*preimage* of b and b is a $\rho$-*image* of a. The first element of pairs in relations is entitled *preimages*. $Dom(\rho) = \{a: (a,b) \in \rho\}$ is the *domain (set)* of $\rho$ ($\rho$ is over $Dom(\rho)$), and $\mathcal{I}(\rho) = \{b: (a,b) \in \rho\}$ is its *image (set)*. Instead of $(a,b) \in \rho$ we often use the notation $a\rho b$. We also say that $\rho$ *is giving b from a*. If the image set for each element of a domain set is a singleton, the concerning binary relation is called a *mapping*. For the relations the postfix notation is the basic presumption ($b = a\rho$); exceptions are relations with some long expressions in domain set or if we want to point out domain elements, and especially for mappings we use prefix notations ($b = \rho a$) or for the sake of clarity $b = \rho(a)$, if needed. We define $\rho: A \mapsto B$, when we want to indicate that $A = Dom(\rho)$ and $B \supseteq \mathcal{I}(\rho)$, and $A\rho B$, if $(a,b) \in \rho$ whenever $a \in A$ and $b \in B$. We also denote $A\rho = \{a\rho : a \in A\}$. When defining mapping $\rho$, we can also use the notation $\rho: a \mapsto b$, $a \in A$ and $b \in B$. If $A \supseteq B$, we say that $\rho$ is a relation in A. When for $\rho: A \mapsto B$ we want to restrict $Dom(\rho)$ to its subset C we denote $\rho_{|C}$, *the restricted mapping of* $\rho$ *to* C for which $\rho_{|C} = \rho \cap \{(c,b): c \in C, b \in B\}$.

Set $\{b: a\rho b\}$ is called the $\rho$-*class* of a. Let $\rho: A \mapsto B$ be a binary relation. We say that $A'(\subseteq A)$ is *closed* under $\rho$, if $A'\rho \subseteq A'$.

For each binary relations $\alpha$, $\chi$ and $\eta$ we define $\alpha(\chi, \eta) = \{(a\chi, b\eta) : (a,b) \in \alpha\}$.

For set $\mathcal{R}$ of relations we denote $a\mathcal{R} = \{ar: r \in \mathcal{R}\}$, $A\mathcal{R} = \{ar: a \in A, r \in \mathcal{R}\}$. If $\rho(A)$ $(=\{\rho(a): a \in A\})$ is B, we call $\rho$ a *surjection*. If $[\rho(x) = \rho(y) \Leftrightarrow x = y\,]$, we call $\rho$ *injection*. If $\rho$ is surjection and injection, we say that it is *bijection*. If $\rho(x) = x$ whenever $x \in Dom(\rho)$, we say that $\rho$ is an *identity* mapping (denoted Id). The element which is an object for the application of a relation is called an *applicant*.

For relations ρ and σ and set $\mathcal{R}$ of relations we define:

the *catenation* $\rho\sigma = \{(a,c): \exists b \in (Dom(\sigma) \cap \mathcal{T}(\rho))\ (a,b) \in \rho,\ (b,c) \in \sigma\}$,

the *inverse* $\rho^{-1} = \{(b,a): (a,b) \in \rho\}$,

$\mathcal{R}^{-1} = \{\rho^{-1}: \rho \in \mathcal{R}\}$.

Let θ be a binary relation in set A. We say that θ is *reflexive*, if $(\forall a \in A)\ (a,a) \in \theta$,

θ is *inversive*, if $\theta^{-1} \subseteq \theta$,

θ is *transitive*, if $\theta\theta \subseteq \theta$,

θ is *associative*, if $(a\theta b)\theta c = a\theta(b\theta c)$,

θ is an *equivalence relation*, if it is reflexive, inversive and transitive. If we want to emphasize the domain, say A, where θ is relation, we denote $\theta \in Eq(A)$.

For sets A and B we define

$|A| = |B|$, if there is such injection α that $\alpha(A)=B$,

$|A| < |B|$, if there is such injection α that $\alpha(A) \subset B$,

$|A| \leq |B|$, if $|A| = |B|$ or $|A| < |B|$.

Set A is *denumerable*, if it is finite or there exists a bijection: $\mathbb{N} \mapsto A$; otherwise it is *undenumerable*.

**Definition 1.1.12.** CARTESIAN POWER. Let $\mathcal{I}$ be a set of index elements and let $\{A_i: i \in \mathcal{I}\}$ be an $\mathcal{I}$-*indexed family* (an index element (shortly index) is incorporated in each element), and let $\mathcal{B}$ be the set of all the bijections joining each set in the indexed family to exactly one element in that set and indexing that element (*an indexed element*) with the index element of that set it includes to. For any element a in $A_i$ we denote $index(a) = i$, and on the other hand for each $i \in \mathcal{I}$, $elem(i,\{A_i: i \in \mathcal{I}\}) = A_i$. Family $\{\{r(A_i): i \in \mathcal{I}\}: r \in \mathcal{B}\}$ (a set of sets consisting indexed elements) is called $|\mathcal{I}|$-*Cartesian power* of indexed family $\{A_i: i \in \mathcal{I}\}$ and we reserve the notation $\Pi(A_i: i \in \mathcal{I})$ for it, and the elements of it are called $|\mathcal{I}|$-*Cartesian elements* on $\{A_i: i \in \mathcal{I}\}$. The cardinality of $\mathcal{I}$, $|\mathcal{I}|$, is called the *Cartesian number* of the elements of $|\mathcal{I}|$-Cartesian power. If $A = A_i$ for each $i \in \mathcal{I}$, we denote $A^{|\mathcal{I}|}$ for $|\mathcal{I}|$-Cartesian power of set A, the elements called $|\mathcal{I}|$-*Cartesian elements of* A. In the case index set $\mathcal{I}$ is $\mathbb{N}$, we denote $(a_1, a_2, \ldots)$ as the element of $|\mathbb{N}|$-Cartesian power of indexed family $A = \{A_i : i \in \mathbb{N}\}$, whenever $a_1 \in A_1$, $a_2 \in A_2, \ldots$ . Any relation from $\mathcal{I}$-Cartesian power to a set is called a $|\mathcal{I}|$-*ary relation*. For the number of Cartesian element $\bar{a}$ we reserve the notation $\mathfrak{N}(\bar{a})$. For finite cases: $n \in \mathbb{N}$ and sets $A_1, A_2, \ldots, A_n$ we define n-*Cartesian power*

$$A_1 \times A_2 \times \ldots \times A_n = \{(a_1, a_2, \ldots, a_n) : a_1 \in A_1, a_2 \in A_2, \ldots, a_n \in A_n\},$$

and call $(a_1, a_2, \ldots, a_n)$ an n-*tuple*. If $\mathcal{I}$ is finite, we can write n-tuple $(a_1, a_2, \ldots, a_n)$, where $|\mathcal{I}| = n$, instead of $\{a_i : i \in \mathcal{I}\}$ and call that tuple the *tuple form* of the $|\mathcal{I}|$-Cartesian element. If $n = 0$, n-Cartesian power is $\emptyset$. Let $\mathcal{K}$ be a set of index elements and let $\{\mathcal{I}_k : k \in \mathcal{K}\}$ be a family of index sets. We denote $\otimes(\Pi(A_i : i \in \mathcal{I}_k) : k \in \mathcal{K}) = \{\{r(A_i) : i \in \mathcal{I}_k, k \in \mathcal{K}\} : r \in \mathcal{B}\}$. For finite $\mathcal{K}$ we can write $\Pi(A_i : i \in \mathcal{I}_1) \otimes \Pi(A_i : i \in \mathcal{I}_2) \otimes \ldots \otimes \Pi(A_i : i \in \mathcal{I}_{|\mathcal{K}|})$ instead of $\otimes(\Pi(A_i : i \in \mathcal{I}_k) : k \in \mathcal{K})$.

For arbitrary Cartesian element $s = \{a_i : i \in \mathcal{I}\}$ we agree on the *undressed notation* of s : $(s) = (a_i | i \in \mathcal{I})$ instead of notation $(\{a_i : i \in \mathcal{I}\})$, and in finite case $(s) = (a_1, a_2, \ldots, a_n)$ instead of $((a_1, a_2, \ldots, a_n))$. For elements $(a_i | i \in \mathcal{I})$ and $(b_j | j \in \mathcal{J})$ $(a_i | i \in \mathcal{I}) = (b_i | i \in \mathcal{I})$, iff $\mathcal{I} = \mathcal{J}$ and for each $(i \in \mathcal{I})$ $a_i = b_i$. We say that relations between two Cartesian powers of indexed families with the same Cartesian number *preserve the indexes*, if in those relations each projection of each preimage and the same projection of its image have the same index.

**Definition 1.1.13.** PROJECTION. Let $\mathcal{I}$ and $\mathcal{J}$ be two arbitrary sets. We call mapping

$e[\mathcal{I}] : (\mathcal{I}, \Pi(A_i : i \in \mathcal{I})) \mapsto \cup(\Pi(A_i : i \in \mathcal{I}))$ a *projection mapping* (reserving that notation for it), where $(\forall j \in \mathcal{I})$ *projection element* $e[\mathcal{I}](j, \bar{a})$ (shortly denoted $\bar{a}_j$) is the element indexed with j in $\bar{a}$ (belonging to $A_j$). We denote simply e, if there is no danger of confusion. We say that a Cartesian element is $\leq$ another Cartesian element, if and only if each projection element of the former is in the set of the projection elements of the latter and the Cartesian number of the former is less than of the latter.

**Definition 1.1.14.** CATENATION. Let $(A_i : i \in \mathcal{I})$ be an indexed set. If each projection element in a $|\mathcal{I}|$-Cartesian element of $\Pi(A_i : i \in \mathcal{I})$ is written before or after another we will get an $|\mathcal{I}|$-*catenation* of family $(A_i : i \in \mathcal{I})$ or a *catenation over $\mathcal{I}$*, and the projections of the concerning Cartesian element are called *members of the catenation*. We denote the set of all $\mathcal{I}$-catenations of family $(A_i : i \in \mathcal{I})$ by $\mathrm{Cat}(A_i : i \in \mathcal{I})$. An associative mapping: $\Pi(A_i : i \in \mathcal{I}) \mapsto \mathrm{Cat}(A_i : i \in \mathcal{I})$ joining an $\mathcal{I}$-*catenation* to each

$\mathfrak{I}$-Cartesian element of $\Pi(A_i: i\in\mathfrak{I})$ is called a *catenation mapping*. Notice that also pq is a catenation, if p and q are catenations, and we say that each member of p *precedes* the members of q and each member of q *succeeds* the members of p; thus preceding and succeeding defining *catenation order* among the members of catenations. If we have a set A such that for each $i\in\mathfrak{I}$ $A_i = A$, we speak of an $|\mathfrak{I}|$-*catenation* of A and denote the set of all the $|\mathfrak{I}|$-catenations of A by $A^{|\mathfrak{I}|}$. E.g. sequence $a_1a_2\ldots a_n$, $n\in\mathbb{N}$, $n > 1$, is a finite catenation. For set H we define $H^*$ (the *catenation closure* of H) such that $H^* = \cup(H^{|\mathfrak{K}|} : \mathfrak{K}\subseteq\mathfrak{I})$. Any catenation of members of catenation c is called a *partial catenation* of c. Such catenation d which are a catenation of partial catenations of catenation c and d = c is called a *decomposition* of c. For our example, above, $d_1d_2$, where $d_1 = a_1a_2\ldots a_i$, $d_2 = a_{i+1}a_{i+2}\ldots a_n$, is a decomposition of $a_1a_2\ldots a_n$. *Catenation operation* $\oplus$ between sets is defined:

$$A\oplus B = \{ab: a\in A, b\in B\}.$$

If the members of a catenation closure are relations we speak of a *transitive closure* of the set of those relations. For set A, index set $\mathfrak{I}$ and set $\mathfrak{R}$ of relations we define:

$$A\mathfrak{R}^{\mathfrak{I}} = (A\mathfrak{R}_i)\mathfrak{R}^{\mathfrak{J}}, \text{ whenever } i\in\mathfrak{I}, \mathfrak{J} = \mathfrak{I}\setminus i \text{ and } \mathfrak{R}_i = \mathfrak{R}.$$

**Definition 1.1.15.** For any symbols x and y we define *replacement* x←y, which means that x is replaced with *substitute* y. Notation A(x←y | C) represents an object where each x occurring in A is replaced with y with condition C; and A(x←∅) is an object where x is deleted.

## 1.2. §     Abstract Data types

### 1.2.1

**Definition 1.2.1.1.** ALPHABETS. Let us introduce three types of alphabets (sets of letters) distinct from each other: *frontier alphabet* (the letters called *variables*, often referred "terminal"), *ranked alphabet* (the letters called *operators*, "nonterminals") Rozenberg G, Salomaa A, ed. (1997) and *arity alphabet* (of *arity* letters) divided in *in-arity* and *out-arity* alphabets, distinct from each other.

The distinction for in- and out-arities is needed in dividing the data flow direction in operations. If there is no danger of confusion symbols X and Y are reserved for frontier alphabets, symbols $\Sigma$ and $\Omega$ are reserved for ranked alphabets and $\Xi$ for the union of in-arity alphabet $\Xi_{in}$ and out-arity alphabet $\Xi_{out}$. The ranked and frontier letters are called *node letters*, shortly nodes, if there is no danger of confusion, and sometimes for frontier letters synonyms leaves are used. Infinite ranks are needed as tools for infinite simultaneous data flow into systems (operations) such as in quantum physics where infinite number of different state function solutions of a Schrödinger-equation compounds a field to be operated.

**Definition 1.2.1.2.** OPERATORS. Let *r* be a mapping from the union of ranked, frontier and arity alphabets to the set of the ordinals assigning to each letter $\gamma$ two ordinals, so called *ranks*, an *in-rank* (in-rank($\gamma$)) and an *out-rank* (out-rank($\gamma$)).

We denote $\Sigma_{\alpha,\beta} = \{\sigma \in \Sigma : \text{in-rank}(\sigma) = \alpha , \text{out-rank}(\sigma) = \beta\}$ to symbolize the set of all $(\alpha,\beta)$-*ary operators* in $\Sigma$. In the special case where the in-rank of an operator of $\Sigma$ is 0 and out-rank = 1, is called *ground letter*. The in-ranks and out-ranks of frontier letters are 1, and the in-ranks and out-ranks of arity letters are 0. In the following the letters and the operators of them are equated with each other, if there is no danger of confusion. Cf. "signatures" (Rozenberg G, Salomaa A, ed. (1997); Nivat M, Reynolds JC, ed. (1985); Ohlebusch E (2002)).

## 1.2.2 Nets

### 1.2.2.1 BASIC DEFINITIONS

Nets describe directed graphs, cf. *model theoretical aspects in consideration of formal descriptive approach for graphs* Thomas W (1997), needed e.g. in computer algorithms and describing connections between neurons in brains, and in more theoretical aspects allowing simultaneous "loops" nets are ideal as formal representations for iterations as e.g. within solutions for powers of higher order differential equations by Picard successive iterants Tirri S, Aurela AM (1989). Without net-formation (differing considerably from trees (Ohlebusch E (2002); Denecke K, Wismat SL (2002)) with only one out-arity) there is no way in a tree to get a return data from any realization of the ranked letter looped to the tree. It is also impossible to cut connection between two parts of one net leaving only subnet and deleting the other part. Also it is impossible to handle simultaneous changes in out-arity connections and furthermore infinite number of out-arities on the whole. Nets allow simultaneous algebraic structures in languages to be recognized by net rewriting automata as would happen in adding the number of saturating term algebra congruence relations by replacing terms with nets and homomorphism relations in tree automata by rewriting systems in Tirri S (1990). Furthermore the results in operation-level in realizations of nets are allowing dependences on coming up streams in carrying nets. By the semantic point of view in process algebra here described nets are concentrating to get in- and output places (filled with arity letters) to ranked letters cf. *tokens* (Best E, Devillers R, Koutny M (2001); Baeten JCM, Basten T (2001)). Some of the preliminary ideas of nets though deviating from knowledge representation, consequently in results, proofs and generalizations are in Tirri SI (2009).

**Definition 1.2.2.1.1.** NETS.

We define $\Sigma X \Xi$-*net* inductively as follows: Each letter in $\Sigma_0 \cup X \cup \Xi$ is a $\Sigma X \Xi$-net. The letters with in-rank 0 are called *ground nets*.

$$t = \sigma(\ r(\xi_i)\ ;\ r(\xi_j)\ |\ i \in \mathcal{I},\ j \in \mathcal{J}\ )$$

is a $\Sigma X \Xi$-*basic net*, the set of its letters $L(t) = \{\sigma\} \cup (\bigcup (L(r(\xi_k)): k \in \mathcal{I} \cup \mathcal{J})$, whenever

(i) $\sigma \in \Sigma$, and $\mathcal{I}$ and $\mathcal{J}$ are such index sets distinct from each other that $|\mathcal{I}|$ = in-rank$(\sigma)$ and $|\mathcal{J}|$ = out-rank$(\sigma)$, and $\{\xi_i : i \in \mathcal{I}\} \subseteq \Xi_{in}$, $\{\xi_j : j \in \mathcal{J}\} \subseteq \Xi_{out}$,

for each $(m, n \in \mathcal{I} \cup \mathcal{J})$ $\xi_m = \xi_n$, iff $m = n$, and

(ii) $\mathcal{I}_o \subseteq \mathcal{I}$, $\mathcal{J}_o \subseteq \mathcal{J}$, and $\mathcal{I}' \subseteq \mathcal{I} \setminus \mathcal{I}_o$, $\mathcal{J}' \subseteq \mathcal{J} \setminus \mathcal{J}_o$,

(iii) for each $(k \in \mathcal{I} \cup \mathcal{J})$ $\alpha_k \in \Sigma$ and $\mathcal{I}_k$ and $\mathcal{J}_k$ are such index sets distinct from each other and from $\mathcal{I}$ and $\mathcal{J}$ that $|\mathcal{I}_k|$ = in-rank$(\alpha_k)$ and $|\mathcal{J}_k|$ = out-rank$(\alpha_k)$, and $\{\xi_m : m \in \mathcal{I}_k\} \subseteq \Xi_{in}$, $\{\xi_n : n \in \mathcal{J}_k\} \subseteq \Xi_{out}$ are such sets of arities that for each $(m, n \in \mathcal{I}_k \cup \mathcal{J}_k)$ $\xi_m = \xi_n$, iff $m = n$, and

(iv) $r$ is such a mapping that

(1.) for each $(k \in \mathcal{I}' \cup \mathcal{J}')$

$$r(\xi_k) = \xi_k\ ,$$

$$L(r(\xi_k)) = \xi_k,$$

where $\xi_k \in \Xi$, and

(2.) for each $(k \in \mathcal{I}_o \cup \mathcal{J}_o)$

$$r(\xi_k) = \xi_k \upsilon_k,$$

$$L(r(\xi_k)) = \{\xi_k, \upsilon_k\},$$

where $\upsilon_k \in X \cup \Sigma_0$, and

(3.) for each $(i \in \mathcal{I} \setminus (\mathcal{I}_o \cup \mathcal{I}'))$ $(\exists\, n_{1_i} \in \mathcal{J}_i)$ $\xi_{n_{1_i}} \in \Xi_{out}$ and

$$r(\xi_i) = \xi_i \xi_{n_{1_i}} \alpha_i(\upsilon_m\,;\,\eta_n\,|\,m \in \mathcal{I}_i,\,n \in \mathcal{J}_i),$$

$$L(r(\xi_i)) = \{\xi_i\} \cup L(\alpha_i(\upsilon_m\,;\,\eta_n\,|\,m \in \mathcal{I}_i,\,n \in \mathcal{J}_i)),$$

where $L(\alpha_i(\upsilon_m\,;\,\eta_n\,|\,m \in \mathcal{I}_i,\,n \in \mathcal{J}_i)) = \{\alpha_i\} \cup \{\upsilon_m, \eta_n : m \in \mathcal{I}_i,\,n \in \mathcal{J}_i\}$,

$\{\upsilon_m : m \in \mathcal{I}_i\} \subseteq \Xi_{in} \cup X \cup \Sigma_0$, $\{\eta_n : n \in \mathcal{J}_i\} \subseteq \Xi_{out} \cup X \cup \Sigma_0$, and

(4.) for each $(j \in \mathcal{J} \setminus (\mathcal{J}_o \cup \mathcal{J}'))$ $(\exists\, m_{0_j} \in \mathcal{I}_j)$ $\xi_{m_{0_j}} \in \Xi_{in}$ and

$$r(\xi_j) = \xi_j \xi_{m_{0_j}} \alpha_j(\upsilon_m, \eta_n\,|\,m \in \mathcal{I}_j,\,n \in \mathcal{J}_j),$$

$$L(r(\xi_j)) = \{\xi_j\} \cup L(\alpha_j(\upsilon_m\,;\,\eta_n\,|\,m \in \mathcal{I}_j,\,n \in \mathcal{J}_j)),$$

where $L(\alpha_j(\upsilon_m\,;\,\eta_n\,|\,m \in \mathcal{I}_j,\,n \in \mathcal{J}_j)) = \{\alpha_j\} \cup \{\upsilon_m, \eta_n : m \in \mathcal{I}_j,\,n \in \mathcal{J}_j\}$,

$\{\upsilon_m : m \in \mathcal{I}_j\} \subseteq \Xi_{in} \cup X \cup \Sigma_0$, $\{\eta_n : n \in \mathcal{J}_j\} \subseteq \Xi_{out} \cup X \cup \Sigma_0$.

We say that for each $(k \in \mathcal{I})$ $r(\xi_k)$ *occupies* arity $\xi_k$ of $\sigma(\upsilon_i\,;\,\eta_j\,|\,i \in \mathcal{I},\,j \in \mathcal{J})$, where $\{\upsilon_i, \eta_j : i \in \mathcal{I},\,j \in \mathcal{J}\} \subseteq \Xi \cup X \cup \Sigma_0$, if $\xi_k \in L(\sigma(\upsilon_i\,;\,\eta_j\,|\,i \in \mathcal{I},\,j \in \mathcal{J}))$ and $r(\xi_k) \in X \cup \Sigma_0$. Furthermore we say that for each $(i \in \mathcal{I} \setminus (\mathcal{I}_o \cup \mathcal{I}'))$ out-arity $\xi_{n_{1_i}}$ of $\alpha_i(\upsilon_m\,;\,\eta_n\,|\,m \in \mathcal{I}_i,\,n \in \mathcal{J}_i)$ and in-arity $\xi_i$ of $\sigma(\xi_i, r(\xi_h)\,;\,r(\xi_k)\,|\,h \in \mathcal{I},\,h \neq i,\,k \in \mathcal{J})$ *occupy each other in* t, and for each $(j \in \mathcal{J} \setminus (\mathcal{J}_o \cup \mathcal{J}'))$ out-arity $\xi_j$ of $\sigma(r(\xi_h)\,;\,\xi_j,\,r(\xi_k)\,|\,h \in \mathcal{I},\,k \in \mathcal{J},\,k \neq j)$ and in-arity $\xi_{m_{0_j}}$ of $\alpha_j(\xi_{m_j},\xi_{n_j}\,|\,m_j \in \mathcal{I}_j,\,n_j \in \mathcal{J}_j)$ occupy each other in t. Furthermore using definitions for symbols defined above for t we define:

$$s = \sigma(\,r(\xi_p),\,\xi_i \xi_{n_i} s_i\,;\,r(\xi_q),\,\xi_j \xi_{m_j} t_j\,|\,p \in \mathcal{I}_o \cup \mathcal{I}',\,q \in \mathcal{J}_o \cup \mathcal{J}',\,i \in \mathcal{I} \setminus (\mathcal{I}_o \cup \mathcal{I}'),\,j \in \mathcal{J} \setminus (\mathcal{J}_o \cup \mathcal{J}'))$$

is a net, and

$$L(s) = \{\sigma,\xi_i,\xi_j : i \in \mathcal{I},\,j \in \mathcal{J}\} \cup (\bigcup(L(r(\xi_k)): k \in \mathcal{I}_o \cup \mathcal{I}' \cup \mathcal{J}_o \cup \mathcal{J}')) \cup (\bigcup(L(s_i): i \in \mathcal{I} \setminus (\mathcal{I}_o \cup \mathcal{I}'))) \cup (\bigcup(L(t_j): j \in \mathcal{J} \setminus (\mathcal{J}_o \cup \mathcal{J}')))$$

is the set of the letters in s,

whenever for each (i∈𝔍\(𝔍₀∪𝔍´), j∈𝔍\(𝔍₀∪𝔍´))

(1.) $s_i$ and $t_j$ are nets outside $\Xi \cup X \cup \Sigma_0$, and

(2.) $\{\xi_n: n \in \mathfrak{I}_i´´\} \subseteq \Xi_{out}$ and $\{\xi_m: m \in \mathfrak{J}_j´´\} \subseteq \Xi_{in}$ are such sets of arities that

for each (m,n∈𝔍ᵢ´´∪𝔍ⱼ´´) $\xi_m = \xi_n$, iff m = n, and

(3.) there is exactly one such index $n_i \in \mathfrak{I}_i´´$ that $\xi_{n_i}$ is an unoccupied out-arity letter in $s_i$, and there is exactly one such index $m_j \in \mathfrak{J}_j´´$ that $\xi_{m_j}$ is an unoccupied in-arity letter in $t_j$.

We call σ the *root* of s, root(s). Net $\sigma(\xi_i; \xi_j | i \in \mathfrak{I}, j \in \mathfrak{J})$ is called the *ranked net*.

For each net u we define and reserve for that purpose such *rank index sets* $\mathfrak{I}_\sigma, \mathfrak{J}_\sigma$, $\sigma \in L(u) \cap (\Sigma \setminus \Sigma_0)$, distinct from each other that $|\mathfrak{I}_\sigma|$ = in-rank(σ) and $|\mathfrak{J}_\sigma|$ = out-rank(σ) and the *in-rank index set of net u* $\mathfrak{I}_u = \bigcup(\mathfrak{I}_\sigma : \sigma \in L(q) \cap (\Sigma \setminus \Sigma_0))$ and the *out-rank index set of net u* $\mathfrak{J}_u = \bigcup(\mathfrak{J}_\sigma : \sigma \in L(q) \cap (\Sigma \setminus \Sigma_0))$.

Uno(u) is a notation for the set of the unoccupied arity letters of u and Occ(u) is reserved for the set of all occupied arity letters of u. Occ(A,t) means the set of all those elements in set A, which are occupied in net t, and Uno(A,t) are reserved for the set of all those which are unoccupied in net t. The index elements in the in-rank index set of net u for unoccupied arities in u are called *unoccupied in-arity index elements*, and the index elements in the out-rank index set of net u for unoccupied arities in u are called *unoccupied out-arity index elements*. The set of all unoccupied in-arity index elements is denoted $\mathfrak{I}_u^{UN}$, and the set of all unoccupied out-arity index elements is denoted $\mathfrak{J}_u^{UN}$. Symbols $\mathfrak{I}_u^{OC}$ and $\mathfrak{J}_u^{OC}$ are reserved for the sets of occupied elements, respectively.

The set of all ΣXΞ-nets is denoted $F_\Sigma(X,\Xi)$. We also denote $F_{\Sigma\Xi}(X) = F_\Sigma(X,\Xi) \setminus \Xi$,

$F_{\Sigma X}(\Xi) = F_\Sigma(X,\Xi) \setminus X$ and $F_{\Sigma X\Xi} = F_\Sigma(X,\Xi) \setminus (X \cup \Xi)$ and $F_{\Sigma X \Sigma_0 \Xi} = F_\Sigma(X,\Xi) \setminus (X \cup \Sigma_0 \cup \Xi)$.

**Definition 1.2.2.1.2.** TIES and SUBNET.

Now when we have reached the definition and the sense of unoccupied arities we are ready to give a formulation for nets in accordance with previous given, more convenient later when we are handling substitutions and rewriting. First we introduce *tied sets* of *tied terms (in tied and out tied) (of nets)*

$\tilde{F}_{in\Sigma}(X,\Xi) = \Xi_{in} \cup \{\xi v : \xi \in \Xi_{in}, v \in X \cup \Sigma_0\} \cup \{\xi_1 \xi_2 u : \xi_1 \in \Xi_{in}, \xi_2 \in \Xi_{out}, \xi_2 \in Uno(u), u \in F_{\Sigma X \Sigma_0 \Xi}\}$ and

$$\widetilde{F}_{out\Sigma}(X,\Xi) = \Xi_{out} \cup \{\xi v : \xi \in \Xi_{out}, v \in X \cup \Sigma_0\} \cup \{\xi_1\xi_2 u : \xi_1 \in \Xi_{out}, \xi_2 \in \Xi_{in}, \xi_2 \in Uno(u), u \in F_{\Sigma X \Sigma_0 \Xi}\}.$$

We denote $\widetilde{F}_{\Sigma}(X,\Xi) = \widetilde{F}_{in\Sigma}(X,\Xi) \cup \widetilde{F}_{out\Sigma}(X,\Xi)$.

The first set and its elements in the union $\widetilde{F}_{in\Sigma}(X,\Xi)$ and $\widetilde{F}_{out\Sigma}(X,\Xi)$ respectively are called *0-tied* and the second set and the elements in it are *1-tied*, and finally the third set and its elements are *2-tied*. For each tied element s we denote its $k^{th}$ member $s^{(k)}$, $k = 1,2,3$. The last member of any tied term is called *tied net* denoted for tied term s by $s_L$, and the preceding members of the tied net are *tie-arities* of $s_L$, the last arity is a *genuine tie-arity* of $s_L$. For 2-tied term s pair $(s^{(1)}, s^{(2)})$ is a 2-*tie* of $s_L$; an *in-2-tie*, if $s \in \widetilde{F}_{in\Sigma}(X,\Xi)$, and an *out-2-tie*, respectively, if $s \in \widetilde{F}_{out\Sigma}(X,\Xi)$. For 1-tied term its first member is the *1-tie* of its last member. 0-tied term is its 0-*tie* itself. For 2-tied term s, $s^{(1)}$ is the 1-*tie* (*in-1-tie*, if $s^{(1)}$ is an in-arity and *out-1-tie*, if $s^{(1)}$ is an out-arity) of $s_L$. We use names in-tied and in-ties and out-tied and out-ties respectively depending on which one of sets $\widetilde{F}_{in\Sigma}(X,\Xi)$ and $\widetilde{F}_{out\Sigma}(X,\Xi)$ those tied elements belong. The set of the in-k-tied and out-k-tied elements ($k = 0,1,2$) are denoted $\widetilde{F}_{in\Sigma}(X,\Xi)^{(k)}$ and $\widetilde{F}_{out\Sigma}(X,\Xi)^{(k)}$ respectively, and the union of those sets by $\widetilde{F}_{\Sigma}(X,\Xi)^{(k)}$. The set of the in-ties in net s is denoted IT(s), and OT(s) for the out-ties, respectively.

For any net s ($\notin X \cup \Xi$) $\xi s$ is named as *an in-gluing form of s*, where $\xi \in \Xi_{in} \cap L(s)$, and if $\xi \in \Xi_{out} \cap L(s)$, $\xi s$ is entitled *an out-gluing form of s*. The set of all in-gluing forms of s is *the in-gluing form of s*, denoted $s_{ing}$ and the set of all out-gluing forms of s is *the out-gluing form of s*, denoted $s_{outg}$. The union $s_{glue} = s_{ing} \cup s_{outg}$ is called *the gluing form of s*, and s is denoted $s_{glueL}$. The gluing form of each letter in $X \cup \Sigma_0$ is the letter itself. The arities have no ranks and therefore either no gluing forms. We define

$\widetilde{F}_{out\Sigma g}(X,\Xi) = \{s_{Lglue} : s \in \widetilde{F}_{out\Sigma}(X,\Xi)\}$,

$\widetilde{F}_{in\Sigma g}(X,\Xi) = \{s_{Lglue} : s \in \widetilde{F}_{in\Sigma}(X,\Xi)\}$ and

$\widetilde{F}_{\Sigma g}(X,\Xi) = \widetilde{F}_{in\Sigma g}(X,\Xi) \cup \widetilde{F}_{out\Sigma g}(X,\Xi)$.

For each $s \in \widetilde{F}_\Sigma(X,\Xi)\}$ we denote $NG(s) = \{s_L, s_{Lglue}\}$. The set of all in-ties in net t is denoted $IT(t)$, and $OT(t)$ for the out-ties, respectively. In-ties and out-ties correspond in-coming and respectively out-coming labeled edges and node letters correspond labeled nodes Engelfriet J (1997).

We define for each $s \in \{\sigma\} \cup F_{\Sigma_{X\Xi}}$

$\quad t = s(\mu_i; \lambda_j \mid i \in \mathcal{I}_s^{UN}, j \in \mathcal{J}_s^{UN}, C)$

is a net, where for each $(i \in \mathcal{I}_s^{UN}, j \in \mathcal{J}_s^{UN})$ $\mu_i \in \widetilde{F}_{in\Sigma}(X,\Xi)$, $\lambda_j \in \widetilde{F}_{out\Sigma}(X,\Xi)$, $\mu_i$ is replacing in-arity letter $\xi_i$ in s and $\lambda_j$ is replacing out-arity letter $\xi_j$ in s, and $\mu_i^{(1)} = \xi_i$, $\lambda_j^{(1)} = \xi_j$, and C is a sample of conditions to be fulfilled (normally assumed to be known) or equivalently

$\quad s(\mu; \lambda \mid C)$

is a net, whenever $\mu \in \widetilde{F}_{in\Sigma}(X,\Xi)^{|\mathcal{I}_s^{UN}|}$, and $\lambda \in \widetilde{F}_{out\Sigma}(X,\Xi)^{|\mathcal{J}_s^{UN}|}$, where the first members in each projection elements of $\mu$ and $\lambda$ are in Uno(s). Nets $\mu_{iL}$, $i \in \mathcal{I}_s^{UN}$, are called *down-subnets of* t, respectively $\lambda_{jL}$, $j \in \mathcal{J}_s^{UN}$, are called *up-subnets of* t, and for each $(q \in sub(t))$ $sub(q) \subseteq sub(t)$, where the set of all subnets of net t is denoted $sub(t)$. Cartesian elements, the projections being nets are *Cartesian nets*. Nets where the out-ranks of the nodes are 1, are *trees*, and trees where the in-ranks of the nodes are 1 are *chains*. We call sets of trees *forests*. A set of nets is called *jungle*, and for jungle T we agree about $sub(T) = \cup(sub(t): t \in T)$. What is said for nets is in the following generalized for jungles as relations from elements to sets of elements and is denoted respectively. E.g. for jungle T we denote $sub(T) = \cup(sub(t): t \in T)$ and $L(T) = \{L(t): t \in T\}$. We say that a *net is finite*, if the cardinalities of the frontier and ranked letters in the net are finite.

Trees can have only nodes with one out-tie at most. The main difference between nets and trees can be demonstrated with the following net containing downstream subnet s incorporating a node with more than one out-tie:

$\quad q(s_i; \lambda_j \mid i \in \mathcal{I}_q^{UN}, j \in \mathcal{J}_q^{UN}, (\forall i \in \kappa) s_{iL} = s, |p(t,s)| = 1, |\{s_i : i \in \kappa\}| > 1)$,

where $\kappa \subseteq \mathcal{I}_q^{UN}$, $p(t,s)$ is defined later in 1.2.2.1.4,

**Definition 1.2.2.1.3.** LINKS and NET CLASSES.

Subnets of nets being frontier letters are called *leaves of the net*, and the set of all leaves in v is denoted by Leav(v). For net v we denote fron(v) as the set of the frontier letters of v, and rank(v) is the set of all ranked letters in v.

For $t = s(\mu_i; \lambda_j \mid i \in \mathcal{I}_s^{UN}, j \in \mathcal{J}_s^{UN}, C)$ net $\mu_{iL}$ is said to be (*next*)*out-linked* to s by out-tie of $\mu_{iL}$, called *out-(arity) linkage* of $\mu_{iL}$, respectively s is said to be (*next*)*in-linked* to $\mu_{iL}$ by in-tie of s, called *in-(arity) linkage* of s. An in- and out-linkage of the same node are said to be *successive* to each other. The linkages between the same two nodes are *parallel* with each other. If net u is out-/in-linked to net q and q is linked to net v, we say that u is *(successively) out-/in-linked to* v. The nets which are not linked to each other are *disjoined* with each other.

A *linkage* (comprising of consecutive in- and out-arity linkages) which connects two nodes in a net is an *inward linkage connection* of the net; the linkages which are not inward connections are *outward linkage connections*. If a net has no outward linkage connections, it is said to be *closed*.

Net $t = \sigma(\mu_i; \lambda_j \mid i \in \mathcal{I}_\sigma, j \in \mathcal{J}_\sigma, C)$ is called $\sigma$-*root revealing net*, where $\sigma \in \Sigma$. Linkages in nets can also be defined by using wider parts of nets: for each $i \in \mathcal{I}_\sigma$ triple $(\sigma, \text{root}(\mu_{iL}), \mu_i^{(1)} \mu_i^{(2)})$ constitutes *node linkage of t*, and $(\text{root}(\mu_{iL}), \sigma, \mu_i^{(2)} \mu_i^{(1)})$ is its *inverse*; respectively for each $j \in \mathcal{J}_\sigma$ $(\sigma, \text{root}(\lambda_{jL}), \lambda_j^{(1)} \lambda_j^{(2)})$ is *node linkage of t*, and $(\text{root}(\lambda_{jL}), \sigma, \lambda_j^{(2)} \lambda_j^{(1)})$ is its *inverse*. The set of the node linkages of t we denote NL(t) and we use notation $\text{NL}^{-1}(t)$ for the set of the inverses of elements in NL(t). Because inverses exist in the up-subnets of root revealing nets, it is natural that "writing directions" of the letters in linkages in nets should not determine those nets. Therefore we will give a sensible definition for the identity of nets:

For nets p and q we define p = q, if $(\forall s \in \text{NL}(p))$ $s \in \text{NL}(q) \cup \text{NL}^{-1}(q)$ and $(\forall s \in \text{NL}(q))$ $s \in \text{NL}(p) \cup \text{NL}^{-1}(p)$. Actually each net defines a *class of nets equal with it* and for net t we denote that set with [t], its elements entitled *t-class representatives*. If there is no danger of confusion, we suppose the appropriate representative to be given.

**Definition 1.2.2.1.4.** POSITION. Next we define locations, *positions*, of nets in nets using arity letters. Let $q = \sigma(s_i; t_j \mid i \in \mathcal{I}_\sigma^{UN}, j \in \mathcal{J}_\sigma^{UN})$ be a net. We say that the out-tie of net $s_{iL}$ in q is a *position* of $s_{iL}$ in q and $s_i^{(1)}$ is the *position* of $s_{iLoutg}$ in q, the sets of the described positions are denoted $p(q, s_{iL})$ ($\subseteq \text{IT}(s_{iL})$), $p(q, s_{iLoutg})$($\subseteq \Xi_{out} \cap L(s_{iLoutg})$), respectively, and $s_{iL}$ and $s_{iLoutg}$ are *next below* q or *next* *down positioned* in q, $i \in \mathcal{S}_\sigma^{UN}$, and an in-tie of $t_{jL}$ in q is a *position* of $t_{jL}$ in q and $t_j^{(1)}$ is the *position* of $t_{jLoutg}$ in q, the corresponding sets denoted $p(q,t_{jL})$ and $p(q,t_{jLoutg})$, and *next above* q or *next up positioned* in q, $j \in \mathcal{S}_\sigma^{UN}$. Furthermore generally for arbitrary nets u and v we define inductively positions as catenations $p(u,v) = p(u,s)p(s,v)$ and $p(u,r) = p(u,s)p(s,r)$, whenever $s \in sub(u)$, $r \in v_{glue}$ and $v \in sub(s)$, next positioned in s. We also say that v and r are *positioned* in u. If c is next above h and h is next above u, we define that c is *above* u. *Below* is defined analogously. The same terminology is a practice also for the positions of the corresponding nets. Next below/next above is denoted shortly by $\leqslant / \geqslant$, and below/above is denoted by $\prec / \succ$. Let $P_1$ and $P_2$ be two arbitrary sets of positions. We define and denote that $P_1 \leqslant P_2$, if $P_1$ and $P_2$ are distinct with each other and $\forall p_1 \in P_1 \; \exists p_2 \in P_2$ such that $p_1 \leqslant p_2$, and $P_1 \prec P_2$, if $\forall p_1 \in P_1 \; p_1 \prec p_2$ whenever $p_2 \in P_2$.

The set of all positioned elements in t is denoted $p(t)$. For sets T and S of nets or gluing forms we denote $p(T,S) = \cup(p(t,s) : t \in T, s \in S)$, and $p(T) = \cup(p(t) : t \in T)$. Furthermore due to the importance of the unoccupied character in nets we take for use notation $Unop(t)$ for the set of the positions of the unoccupied arities in t, and generalize the notation as usual for jungle, say T, $Unop(T) = \cup(Unop(t) : t \in T)$. Furthermore for jungle T we denote the cardinality of $Unop(T)$ by $\delta_D(T)$. Cf. "marked letters" Ohlebusch E (2002).

For net v, v|p (an *occurrence*), is denoted to be the subnet of v having or "topped at" position p in v. A *down-/up-frontier net* of net v, down-/up-fronnet(v), is such a subnet of v, whose occurrence is next below/next above v (at so called down-/up-*frontier position* of v). We denote Frd(v) meaning the set of all down-frontier nets of v, and Fru(v) is the set of all up-frontier nets of v, and $F_r(v)$ means the set of all frontier nets of v.

We define the height of net t, hg(t), by the following induction:

1°  $hg(t) = 0$, if $t \in \Xi \cup X \cup \Sigma_0$

2°  $hg(t) = 1 + \max\{hg(s) : s \in F_r(t)\}$, if $t \in F_\Sigma(X,\Xi) \setminus (\Xi \cup X \cup \Sigma_0)$.

For arbitrary net t, there is in force equation $|[t]| = |\{p(t,\sigma) : \sigma \in L(t) \cap \Sigma\}|$.

Notice that for any net t and its subnet s outside $X \cup \Sigma_0$, the positions of s and its gluing form in t are different and that the position of s is unequivocal, but its gluing form can be rearrange in many ways to the context of t next to it (thus forming new nets), depending on which arities of the gluing form is chosen to occupy arities of the context. Trees (owing only one out-arity) have naturally no such difference between the positions of nets and the gluing form of them. We will come to this matter of rearrangement more profoundly later in the chapter of rewriting.

**Definition 1.2.2.1.5.** ENCLOSEMENTS. Let $t = s(\mu_i; \lambda_j \mid i \in \mathcal{I}_s^{UN}, j \in \mathcal{J}_s^{UN})$ be a net. We call out-gluing forms $\mu_{iLoutg}$, $i \in \mathcal{I}_s^{UN}$, s-*downstream elements* of t for s (s being the *upcontext of t for those elements, or for the set of them*) and in-gluing forms $\lambda_{jLing}$, $j \in \mathcal{J}_s^{UN}$, s-*upstream elements* of t for s (s being the *down context of t for those elements*). If we want to emphasize that net v is the context of net u only for the frontier letters in u, we say that v is the *apex of* u, apex(u), for those letters; accordingly *down and up apex*, respectively. We agree of notation apex(T) = {apex(t): t∈T}, whenever T is a jungle.

Net s can also be expressed with notation $con_P(t)$, where

$P = \{ p(t, \mu_{iLoutg}), p(t, \lambda_{jLing}) : i \in \mathcal{I}_s^{UN}, j \in \mathcal{J}_s^{UN} \}$.

Notice that context $con_P(t)$ is the apex of t, if $P = \{p(t,x) : x \in X \cap L(s)\}$.

The sets of the ties of $\mu_{iL}$ and $\lambda_{jL}$ to s, $i \in \mathcal{I}_s^{UN}$, $j \in \mathcal{J}_s^{UN}$, are *matching arity linkage sets of s to t* and the family of all of them is denoted MAL(t,s). We also call s the *abover* of $\mu_{iL}$ in t, denoted $t\backslash_b\{\mu_{iL}: i \in \mathcal{I}_s^{UN}\}$, and each $\mu_{iL}$, $i \in \mathcal{I}_s^{UN}$ is a *belower* of s in t, the set of the belowers of context s in t is denoted $t\backslash_a s$.

We say that *net s is linked outside to net t*, if s is linked to t and the set of the arity linkages of s to t differs from set MAL(t,s).

If u is a subnet of net v, we say that v can be *divided in two nets* : u and the abover of u in v. The contexts of the subnets of t are the *enclosures* of t (we say they are in t or t is embedding them), and the set of all enclosures of t is denoted enc(t). For jungle T we denote enc(T) = ∪(enc(t) : t∈T). We say that a *net is finite*, if the cardinalities of the frontier and ranked letters in the net are finite. The elements in enc(t)\t are entitled *genuine enclosures of t* and the set of them is denoted $enc_g(t)$; for jungle T we have $enc_g(T) = \cup(enc_g(t) : t \in T)$.

For further need it is worth to notice that because for any net t NL(t) = ∪(NL(s) : s∈enc(t)), we can write [enc(t)] = enc([t]).

**Definition 1.2.2.1.6.** OVERLAPPING and OMISSION.

Let p and q be arbitrary nets. If there is such net t, each enclosement of which is in both [p] and [q] and has a linkage connection to each other, we say that p and q *overlap* each other, and t is said to be *shared* among p and q. If $E^{pq}$ is the denotation for the set of all shared nets among p and q, *the overlapping net of p and q* denoted p⋒q is such a net in $E^{pq}$ (the most "extensive") that $(\forall k \in E^{pq})$ k∈enc([p⋒q]). For jungles P and Q we define P⋒Q is such a net in $E^{pq}$ (= $\cap(E^{pq} : p \in P, q \in Q)$) that k∈enc([p⋒q]) whenever k∈$E^{pq}$, and ⋒Q = Q⋒Q. Nets are said to be *distinctive from each other*, if they do not overlap each other. A *jungle is distinctive* if all of its nets are distinctive from each other, and furthermore a relation over a distinctive jungle domain is entitled a *distinctive relation*.

For an arbitrary nets s and t *the set of the positions of the outside arities of* t *in* s, (Unop(t,s)), means the set of the positions of all those arities of the elements in L(t⋒s) which are not occupied by anything in net s.

Let s and t be two arbitrary nets. Let $s^o$ be the context of such a representative of [s] that the context is for the gluing form of a representative of [s⋒t], and respectively let $t^o$ be the context of such a representative of [t] that the context is for the gluing form of a representative of [s⋒t]. Jungle {$s^o, t^o$} is called the *omission of s by t or s omitted by t*, denoted s⋌t. Notice that an omission may be broken (cf. "broken jungle" defined later). For arbitrary net s and jungle S we denote s⋌T = ∩(s⋌t : t∈T) and for jungles S and T we use notation S-T = {s⋌T : s∈S}.

Net, say k, possessing nets s and t as subnets and for which k⋌t = s⋌t and k⋌s = t⋌s is *the assimilation* of s and t and we denote s⋓t.

## 1.2.2.2 Characters of Nets

Here we represent some features typical to nets and the relations between them.

**Definition 1.2.2.2.1.** NEIGHBOURING, ISOLATION and BORDER.

If nets do not overlap each other, but are linked to each other, we say they are *neighbouring* each other. A set of the neighbouring nets of a net is called a *touching surrounding of the net*. Nets are said to be *isolated* from each other, if there is a net neighboured by them. We say that nets being neighboured by each other are *linked directly*, and nets being isolated from each other are *linked via isolation*.

If nets are neighbouring each other such that they are not isolated from each other, we say they are *closely neighbouring* each other.

If nets are isolated from each other, but are not neighbouring each other, we say they are *totally isolated* from each other.

Net s is t-*isolated*, if the nodes of t are totally isolated from each other by the nodes of s, and inversely.

The set of the linkages connecting two nets to each other is called the *border* between those nets. The border may be empty, too. The union of the set of the borders between a net and all other neighbouring nets is called simply *the border of the net*.

**Definition 1.2.2.2.2.** THE RIM and BROKEN JUNGLE.

The nets of a jungle which are in-linked inside the jungle, but not out-linked, are *out-end nets* and at *out-end positions* in the jungle, and the nets out-linked inside a jungle, but not in-linked, are *in-end nets* and at *in-end positions* in the jungle. The union of the in-end nets and the out-end nets in a jungle is the *rim of the jungle*.

We call a jungle *broken*, if each of its nets is disjoined from each other; otherwise it is *unbroken*. Notice that unbroken jungles are actually nets. Broken jungles, each net having only one letter outside the arity alphabet, are *totally broken*. E.g. any set, the elements of which are nodes, can be seen as a totally broken jungle and is called *degenerated*. Because of the close relationship between nets and jungles we often denote jungle by small letter instead of the normal procedure for sets. Comparative study in the form of *dependence* can be found in Diekert V, Métivier Y (1997).

Notice that even if a net itself is unbroken, an enclosement of it may be a broken jungle.

**Definition 1.2.2.2.3.** ROUTES and LOOPS.

A *denumerable route* (DR) between nets is defined as follows:

1°  any linkage between two nets is a route between those nets, and
2°  if P is a DR between net s and t, and Q is a DR between t and net u, then PQ is a DR between s and u.

DR can also be seen as an inversive and transitive relation in the set of the nets, if "linkage" is interpreted as a binary relation in the set of the nets. Any route can also be denoted by the catenation of the nets linked with each other in the route. Cf. *paths* and *cycles* Müller J (1997).

We define an *in-/out-one-way* DR (in-/out-OWR) between nets as transitive relation ("linkage" is a binary relation) among the set of the nets as follows:

1°  any linkage which is an in-/out-linkage of net s and on the other hand an out-/in-linkage of net t is an in-/out-OWR from s to t, and
2°  if P is an in-/out-OWR from net s to net t, and Q is an in-/out-OWR from t to net u, then PQ is an in-/out-OWR from s to u, and we say that s *in-/out-dominates* u and u *out-/in-dominates* s.

Nets s and t are A- or |A|-*routed* with each other, if A is the set of routes between them. Cf. *Trace semantic* (van Glabbeek RJ (2001); Aceto L, Fokkink WJ, Verhoef C (2001)).

Triple ($\mathcal{N},\mathcal{R},f$), where $\mathcal{N}$ is a jungle, $\mathcal{R}$ is a set of OWR´s and f is a mapping connecting the elements of $\mathcal{R}$ to pairs of nets, describes *graph* (Rozenberg G, Salomaa, A ed. (1997); Müller J (1997)).

If there is no need to distinguish in- and out-arities from each other we write nets simply compounding indexes for in- and out-arities to be one index and interpret out-arities as in-arities.

An DR from a net to itself is a *loop* of the net, and *outside* loop, if furthermore in the route there is a linkage to outside the net; otherwise it is an *inside loop of the net*. The loop where each linkage is among the linkages of the same jungle, is an *inside loop of the jungle*. OWR´s which are loops (*OWR-* *loops*) are called *directed loops*. A *bush* is a jungle which has no inside loops and *elementary*, if it has no parallel linkages between its nets.

The following is an example of equal nets, containing an inside directed loop; $\sigma$, $\rho$, $\lambda$ and $\mu$ are nets, not including to arities ($\xi$ stands for in-arity, $\bar{\xi}$ for out-arity) nor frontier letters:

$s = t(\xi_t \bar{\xi}_{v1} \; v(\xi_{v1}\xi_\mu \; \mu, \xi_{v2} \; \bar{\xi}_{u2}q \; ; \; \bar{\xi}_{v2}\xi_\lambda \lambda) \; ; \; \bar{\xi}_t \xi_{u2} u(\xi_{u1}\xi_\rho \rho, \; \xi_{u2}\xi_t \; s \; ; \; \bar{\xi}_{u1}\xi_\sigma \sigma, \; \bar{\xi}_{u2}\xi_{v2}r))$

$q = u(\xi_{u1}\xi_\rho \rho, \xi_{u2} \; \bar{\xi}_t s \; ; \; \bar{\xi}_{u1}\xi_\sigma \sigma, \; \bar{\xi}_{u2}\xi_{v2} \; r)$

$r = v(\xi_{v1}\xi_\mu \; \mu, \xi_{v2} \; \bar{\xi}_{u2}q \; ; \; \bar{\xi}_{v1}\xi_t s, \; \bar{\xi}_{v2}\xi_\lambda \lambda)$

$enc_g(s) = enc(\{u,\rho,\sigma,r,q,\mu,\lambda,t,v\})$
$enc_g(q) = enc(\{u,\rho,\sigma,r,s\})$
$enc_g(r) = enc(\{\mu,\lambda,v,s,q\})$

This yields s,q and r are enclosements of each other and we have s = q = r.

### 1.2.3 Realizations, Algebras and Homomorphisms

In this paragraph we introduce the notions of nets in semantic point of view referring to algebras overall. We represent generalization for more common $\Sigma$-algebra definition (Aceto L, Fokkink WJ, Verhoef C (2001); Burris S, Sankappanavar HP (1981)) – concerning nets. Net rewriting, represented closely later, can be guided by realizations of nets which realizations can be understood also to correspond on temporal logic and models Gabbay DM, Hogger CJ, Robinson JA (1995). Due to the generic essentiality relations between algebras in the respect of free generation and morphisms are briefly taken into the consideration.

**Definition 1.2.3.1.** OPERATIONS and $\Sigma X\Xi$-ALGEBRA. The represented definition for nets allows upstream subnets of nets to influence in producing the images of realizations of the roots of the nets. An (upstream related or *look-ahead*) $\Sigma_{\mathcal{A}}X_{\mathcal{A}}\Xi_{\mathcal{A}}$-*algebra* $\mathcal{A}$, $\Sigma_{\mathcal{A}} \subseteq \Sigma$, $X_{\mathcal{A}} \subseteq X$, $\Xi_{\mathcal{A}} \subseteq \Xi$, is a pair consisting of a set A (including $\Xi_{\mathcal{A}in}$), and a mapping, an *operation assigning mapping*, that assigns to each operator of $\Sigma_{\mathcal{A}} \cup X_{\mathcal{A}} \cup \Xi_{\mathcal{A}}$ ($\Sigma_{\mathcal{A}} \subseteq \Sigma$, $X_{\mathcal{A}} \subseteq X$, $\Xi_{\mathcal{A}} \subseteq \Xi$) $\mathcal{A}$-*operation*, to $(\alpha,\beta)$-ary operator $\sigma \in \Sigma_{\mathcal{A}}$ a relation, $(\alpha,\beta)$-*ary operation* $\sigma^{\mathcal{A}} : A^{\alpha} \otimes \widetilde{F}_{out\Sigma}(X,\Xi)^{\beta} \mapsto A$, where $\alpha$ = in-rank($\sigma$) and $\beta$ = out-rank($\sigma$), and to each letter $x \in X_{\mathcal{A}}$ *operation* $x^{\mathcal{A}}$ for which $x^{\mathcal{A}}(a,t) = a$, whenever $a \in A$, $t \in \widetilde{F}_{out\Sigma}(X,\Xi)$, and to each arity letter $\xi \in \Xi_{\mathcal{A}in}$ *constant operation* $\xi^{\mathcal{A}}$ in $\Xi_{\mathcal{A}in}$. The operations of the ground letters are defined by the constant images in A. For simplicity we write $\mathcal{A} = (A,\Sigma_{\mathcal{A}}X_{\mathcal{A}}\Xi_{\mathcal{A}})$ and assume the operation assigning mapping to be known. We say that $\mathcal{B} = (B,\Sigma_{\mathcal{B}}X_{\mathcal{B}}\Xi_{\mathcal{B}})$, where B (including $\Xi_{\mathcal{B}in}$) is a subset of A, is a *subalgebra* of $\mathcal{A}$, if $\sigma^{\mathcal{B}} = \sigma^{\mathcal{A}}|B^{\alpha} \otimes \widetilde{F}_{out\Sigma}(X,\Xi)^{\beta}$, and the image set of $\sigma^{\mathcal{B}}$ is in B, whenever $\sigma \in \Sigma_{\mathcal{B}}$, and $x^{\mathcal{B}} = x^{\mathcal{A}}|B$ and $\xi^{\mathcal{B}} = \xi^{\mathcal{A}}|B$. Set B is called a *closed subset of* A. Sub($\mathcal{A}$) symbolizes the set of all subalgebras of $\mathcal{A}$. It is worth to mention that up-subnets in realizations of root revealing nets are important as e.g. in TD (defined later), where images of operations may in that way be selected to direct to desired in-arities. For each $\Sigma_{\mathcal{A}}X_{\mathcal{A}}\Xi_{\mathcal{A}}$-algebra $\mathcal{A} = (A,\Sigma_{\mathcal{A}} \cup X_{\mathcal{A}} \cup \Xi_{\mathcal{A}})$ we define *the power algebra of* $\mathcal{A}$, $P(\mathcal{A}) = (P(A),\Sigma_{P(\mathcal{A})} \cup X_{P(\mathcal{A})} \cup \Xi_{P(\mathcal{A})})$, where for each of its element A´ and operation $\sigma^{P(\mathcal{A})}$, $A´\sigma^{P(\mathcal{A})} = \{a\sigma^{\mathcal{A}} : a \in A´\}$.

**Definition 1.2.3.2.** $\Sigma X \Xi$-NET ALGEBRA. Algebra $\mathcal{F}_{\Sigma}^{\Xi}(X) = (F_{\Sigma}(X,\Xi), \Sigma X \Xi)$ defined so that for each operator $\sigma \in \Sigma$ and $\Sigma X \Xi$-nets $s_i$, $i \in \mathcal{I}_{\sigma}$, and $\lambda_j \in \widetilde{F}_{out\Sigma}(X,\Xi)$, $j \in \mathcal{J}_{\sigma}$

$$\sigma^{\mathcal{F}_{\Sigma}^{\Xi}(X)}(s_i;\lambda_j | i \in \mathcal{I}_{\sigma}, j \in \mathcal{J}_{\sigma}) = \sigma(\mu_i;\lambda_j | i \in \mathcal{I}_{\sigma}, j \in \mathcal{J}_{\sigma}), \text{ if } \sigma \notin \Sigma_0, \text{ and}$$

$$\sigma^{\mathcal{F}_{\Sigma}^{\Xi}(X)}(s_i;\lambda_j | i \in \mathcal{I}_{\sigma}, j \in \mathcal{J}_{\sigma}) = \sigma, \text{ if } \sigma \in \Sigma_0,$$

whenever for each $i \in \mathcal{I}_{\sigma}$, $\mu_i \in \widetilde{F}_{in\Sigma}(X,\Xi)$ and $\mu_{iL} = s_i$, and

$$\gamma^{\mathcal{F}_{\Sigma}^{\Xi}(X)}(s) = s \quad \text{for each } s \in \Sigma_0 \cup X \cup \Xi \text{ and } \gamma \in X \cup \Xi_{in},$$

is called the $\Sigma X \Xi$-*net algebra* or *free algebra*. If in the $\Sigma X \Xi$-net algebra we interchange in each ranked letter the in-arities and out-arities we will get *the co-algebra of* the $\Sigma X \Xi$-net algebra.

**Definition 1.2.3.3.** REALIZATION of NETS. Operation $\gamma^{\mathcal{A}}$ is $\mathcal{A}$-*realization* of $\gamma$, whenever $\gamma \in \Sigma_0 \cup X \cup \Xi$. Let $t = s(\mu_i; \lambda_j \mid i \in \mathcal{I}_s^{UN}, j \in \mathcal{J}_s^{UN})$ be a net. Then

$$t^{\mathcal{A}} = s^{\mathcal{A}}(\mu_{iL}^{\mathcal{A}}, \lambda_j \mid i \in \mathcal{I}_s^{UN}, j \in \mathcal{J}_s^{UN}, C_{t\mathcal{A}})$$

is $\mathcal{A}$-*realization* of t, where $C_{t\mathcal{A}}$ is a sample of conditions to be fulfilled (normally assumed to be known). If net t is given in the form $t = s(\mu; \lambda)$, then we can write $\mathcal{A}$-realization of t

$$t^{\mathcal{A}} = s^{\mathcal{A}}(\mu_L^{\mathcal{A}}, \lambda),$$

when $\mu_L^{\mathcal{A}}$ is a Cartesian element where each projection of $\mu_L$ is replaced with its $\mathcal{A}$-realizations. For jungle T we denote $T^{\mathcal{A}} = \{t^{\mathcal{A}} : t \in T\}$, and the set of $\mathcal{A}$-realizations of all $\Sigma X\Xi$-nets is denoted $F_\Sigma(X,\Xi)^{\mathcal{A}}$.

Net s is called the *carrier net* of $s^{\mathcal{A}}$. Let $\mathcal{A} = (A, \Sigma_{\mathcal{A}} X_{\mathcal{A}} \Xi_{\mathcal{A}})$ be a $\Sigma_{\mathcal{A}} X_{\mathcal{A}} \Xi_{\mathcal{A}}$-algebra. We may also use notation $t = (t^{\mathcal{A}})^{-\mathcal{A}}$, and for jungle S, $S = (S^{\mathcal{A}})^{-\mathcal{A}}$. The set of the $\mathcal{A}$-operations of the nodes in t is entitled $\mathcal{A}$-*nest* of t or the nest of $t^{\mathcal{A}}$, t and $t^{\mathcal{A}}$ being said to be *beyond* any subset of that nest.

Let $t = s(\mu_i; \lambda_j \mid i \in \mathcal{I}_s^{OC(X)}, j \in \mathcal{J}_s^{UN})$ be a net, where $\mathcal{I}_s^{OC(X)}$ is such an in-rank index set of s that for each $(i \in \mathcal{I}_s^{OC(X)})$ $\mu_i \in \widetilde{F}_{in\Sigma}(X,\Xi)^{(1)}$. Let $A_o = \{a_i : a_i \in A, i \in \mathcal{I}_s^{OC(X)}\}$ be an indexed subset of A, *a set of inputs*.

$$t^{\mathcal{A}}(A_o) = s^{\mathcal{A}}(\mu_{iL}^{\mathcal{A}}(a_i), \lambda_j \mid i \in \mathcal{I}_s^{OC(X)}, j \in \mathcal{J}_s^{UN}, a_i \in A_o, C_{t\mathcal{A}})$$

is called $t^{\mathcal{A}}$-*transformation of* $A_o$, *the set of outputs of* $t^{\mathcal{A}}$ *for* $A_o$. Important examples of realizations are equations, where e.g. symbol "=" is the realization of a ranked letter with the in-rank two, and transformations are needed to considering the validity.

Transformations are essential in the context of rewriting systems.

Let $\mathfrak{r}$ be a binary relation in P(A). $\mathcal{A}$-realization $t^{\mathcal{A}}$ is $\mathfrak{r}$-*confluent*, if $t^{\mathcal{A}}(A) \mathfrak{r} t^{\mathcal{A}}(B)$, whenever $A\mathfrak{r}B$.

Net realization descriptions

**Lemma 1.2.3.** Each demand or claim can always be presented with realizations of nets.

PROOF. Each presentable elementary claim is actually a relation in some algebra. □

**Definition 1.2.3.4.** Let $\mathcal{A} = (A, \Sigma X \Xi)$ be a $\Sigma X \Xi$-algebra. Let R, S and T be $\mathcal{A}$-realizations of some nets. Now we are introducing for only descriptive use some special nets by example wise: *Transformer graph* (TG) $\mathcal{T}$ over {R,S,T}, denoted TG({R,S,T}), is a realization of the net having carrier nets of R, S and T among its enclosements. If the set, for which a transformer graph is over, is singleton, we speak simply of *transformer*. If H is a set of realizations, set K being one of the subsets of H, we say that $\mathcal{T}$ is *beyond* K whenever $\mathcal{T}$ is TG(H) and we denote TG(|K).

*Realization process graph* (RPG) contains in an addition to TG input and output elements as its nodes for concerning realizations. Cf. *model theory* Chang CC, Keisler HJ (1973), *process graphs* van Glabbeek RJ (2001), *automata* Rozenberg G, Salomaa A, ed. (1997).

Generally speaking: any RPG is a TG-associated net, where the projections of Cartesian elements of A in the RPG are in- and up-connected, respectively, to at most one $\mathcal{A}$-realization in the TG.

*Transformation graph* (TFG) comprises only input and output elements of RPG.

**Definition 1.2.3.5.** GENERATORS. Let $\mathcal{A} = (A, \Sigma_{\mathcal{A}} X_{\mathcal{A}} \Xi_{\mathcal{A}})$ be an algebra. We say that subset H of A is *a generator set of* $\mathcal{A}$ and is *generating* $\mathcal{A}$, and we denote [H] = A, if

$A = \bigcap ( B : H \subseteq B, (B, \Sigma_{\mathcal{B}} X_{\mathcal{B}} \Xi_{\mathcal{B}}) \in \text{Sub}(\mathcal{A}))$.

H is called *a base generator set* of $\mathcal{A}$, if there is no genuine subset of H generating $\mathcal{A}$. Notice that algebra may have several base generator sets.

**Definition 1.2.3.6.** HOMOMORPHISM. Let $\mathcal{A} = (A, \Sigma_{\mathcal{A}} X_{\mathcal{A}} \Xi_{\mathcal{A}})$ and $\mathcal{B} = (B, \Sigma_{\mathcal{B}} X_{\mathcal{B}} \Xi_{\mathcal{B}})$ be two algebras. Let $\varphi : A \mapsto B$ be an indexes preserving relation. *The homomorphic extension of* $\varphi$ *from* $\mathcal{A}$ *to* $\mathcal{B}$, shortly *homomorphism*, is a relation, denoted $\hat{\varphi}: \mathcal{A} \mapsto \mathcal{B}$, defined such that

$\hat{\varphi}(a) = \varphi(a)$, whenever $a \in A$, and

$\hat{\varphi}(\sigma^{\mathcal{A}}(a_i; \lambda_j \mid i \in \mathcal{I}_\sigma, j \in \mathcal{J}_\sigma)) = \sigma^{\mathcal{B}}(\hat{\varphi}(a_i); \lambda_j \mid : i \in \mathcal{I}_\sigma, j \in \mathcal{J}_\sigma)$,

whenever $\sigma\in\Sigma$ and $(a_i;\lambda_j \mid i\in\mathcal{I}_\sigma, j\in\mathcal{J}_\sigma) \in A^{|\mathcal{I}_\sigma|} \otimes \widetilde{F}_{out\Sigma}(X,\Xi)^{|\mathcal{J}_\sigma|}$. Homomorphism : $\mathcal{A} \mapsto \mathcal{A}$ is *automorphism*.

**Definition 1.2.3.7.** FREE GENERATING. Let K the set of all $\Sigma X\Xi$-algebras. We say that *free generating set* $A_o$ *generates* $\mathcal{A} = (A,\Sigma)$ *freely over* K, if there is such subset $A_o\subseteq A$ that

(i)  $[A_o] = A$ and

(ii) for each algebra $\mathcal{B} = (B,\Sigma)$ in K and each relation $\varphi:A_o\mapsto B$, there is the homomorphic extension of $\varphi$ from $\mathcal{A}$ to $\mathcal{B}$.

The next clause represents the well known result for trees, at this time for more general aspects: nets.

**Proposition 1.2.3.** Set $\Xi\cup X\cup\Sigma_0$ is the base generator set of $\Sigma X\Xi$-net algebra $\mathcal{F}_\Sigma^\Xi(X)$ and generates it freely over all $\Sigma X\Xi$-algebras.

PROOF.   First we prove that $\Xi\cup X\cup\Sigma_0$ is the base generator set of $\mathcal{F}_\Sigma^\Xi(X)$ and for that purpose we define for each jungle $Q\subseteq F_\Sigma(X,\Xi)$ the following sets:

$G(Q) = Q\cup\{\sigma^{\mathcal{F}_\Sigma^\Xi(X)}(s_i;\lambda_j \mid i\in\mathcal{I}_\sigma, j\in\mathcal{J}_\sigma) : s_i\in Q, \lambda_j\in\{\lambda : \lambda\in\widetilde{F}_{out\Sigma}(X,\Xi), \lambda_L\in Q\}, \sigma\in\Sigma\setminus\Sigma_0\}$,

$G^0(Q) = Q$,

$G^{n+1}(Q) = G(G^n(Q))$, $n\in \mathbb{N}_0$,

$G = \cup( G^n(Q) : n\in\mathbb{N}_0 )$.

Let $\sigma\in\Sigma$ and $\mathcal{I}'\subseteq\mathcal{I}_\sigma$ and $\mathcal{J}'\subseteq\mathcal{J}_\sigma$, and $s_i,t_j \in G$, $i\in\mathcal{I}'$, $j\in\mathcal{J}'$. Then ( $\forall$ $i\in\mathcal{I}'$, $j\in\mathcal{J}'$ ) ( $\exists$ $m_i,n_j\in \mathbb{N}_0$ ) $s_i\in G^{m_i}(Q)$ and $t_j\in G^{n_j}(Q)$. Therefore

$s_i\in G^m(Q)$, $i\in\mathcal{I}'$ and $t_j\in G^m(Q)$, $j\in\mathcal{J}'$,

where m is the maximum of the numbers $n_k$ , $k\in\mathcal{I}'\cup\mathcal{J}'$. We can write

$\sigma^{\mathcal{F}_\Sigma^\Xi(X)}(s_i;\lambda_j \mid i\in\mathcal{I}_\sigma, j\in\mathcal{J}_\sigma) \in G(G^m(Q)) = G^{m+1}(Q) \subseteq G$, where $\lambda_{jL}=t_j$.

This yields G generates itself. Because $Q\subseteq G$, it is in force

[Q] ⊆ G.

On the other hand for each ($n \in \mathbb{N}_0$) $G^n(Q) \subseteq [Q]$. Therefore $G \subseteq [Q]$. If $Q = \Xi \cup X \cup \Sigma_0$, we thus have $G = F_\Sigma(X, \Xi)$, and finally $[\Xi \cup X \cup \Sigma_0] = F_\Sigma(X, \Xi)$.

Let then $\mathcal{A} = (A, \Sigma_\mathcal{A}, X_\mathcal{A}, \Xi_\mathcal{A})$ be an $\Sigma X \Xi$-algebra and $\varphi: \Xi \cup X \cup \Sigma_0 \mapsto A$ an indexes preserving relation. We define relation $\beta: \mathcal{F}_\Sigma^\Xi(X) \mapsto \mathcal{A}$ such that

$\beta(\gamma) = \varphi(\gamma)$, whenever $\gamma \in \Xi \cup X \cup \Sigma_0$, and

$\beta(\sigma^{\mathcal{F}_\Sigma^\Xi(X)}(s_i; \lambda_j \mid i \in \mathcal{I}_\sigma, j \in \mathcal{J}_\sigma)) = \sigma^\mathcal{A}(\beta(s_i); \lambda_j : i \in \mathcal{I}_\sigma, j \in \mathcal{J}_\sigma)$,

whenever $\sigma \in \Sigma$ and $(s_i; \lambda_j \mid i \in \mathcal{I}_\sigma, j \in \mathcal{J}_\sigma) \in A^{|\mathcal{I}_\sigma|} \otimes \widetilde{F}_{out\Sigma}(X,\Xi)^{|\mathcal{J}_\sigma|}$. Clearly $\beta$ is the homomorphic extension of $\varphi$ from $\mathcal{F}_\Sigma^\Xi(X)$ to $\mathcal{A}$. □

## 1.3. § Net homomorphism, Substitution and Matching

**Definition 1.3.1.** NET HOMOMORPHISM.

Let X and Y be frontier alphabets, $\Sigma$ and $\Omega$ ranked alphabets and $\Xi_\Sigma$ and $\Xi_\Omega$ arity alphabets. We introduce new distinct rank-indexed arity alphabets $E_{in} = \{\varepsilon_i : i \in \mathcal{E}_{in}\}$ for in-arities and $E_{out} = \{\varepsilon_i : i \in \mathcal{E}_{out}\}$ for out-arities respectively, disjoint from all other used alphabets.

*Net homomorphism* h: $F_\Sigma(X, \Xi_\Sigma) \cup \widetilde{F}_\Sigma(X, \Xi_\Sigma) \mapsto F_\Omega(Y, \Xi_\Omega) \cup \widetilde{F}_\Omega(Y, \Xi_\Omega)$ is a relation defined such that

$h(t) = h_\Sigma(\sigma)(h(\mu_i); h(\lambda_j) \mid i \in \mathcal{E}_{inh_\Sigma(\sigma)}, j \in \mathcal{E}_{outh_\Sigma(\sigma)})$ for each $t = \sigma(\mu_i; \lambda_j \mid i \in \mathcal{I}_\sigma, j \in \mathcal{J}_\sigma) \in F_\Sigma(X, \Xi_\Sigma)$,

and

h: $\Sigma_0 \cup X \cup \Xi_\Sigma \mapsto \widetilde{F}_\Omega(Y, \Xi_\Omega)^{(1)} \cup \Omega_0 \cup \Xi_\Omega$ is an *initial rewriting relation*, where $h(\xi) \in \Xi_\Omega$ for each $\xi \in \Xi_\Sigma$ and $h(\sigma) \in \Omega_0$ whenever $\sigma \in \Sigma_0$;

h|X named the *initial manoeuvre rewriting relation*, and h|$\Xi_\Sigma$ the *initial arity rewriting relation*;

$h_\Sigma: \Sigma \mapsto F_\Omega(Y, \Xi_\Omega \cup E_{in} \cup E_{out}) \cup \Omega$ is a $\Sigma$-*ranked letter rewriting relation*,

$h(u) \in \widetilde{F}_{in\Omega}(Y, \Xi_\Omega)^{(1)}$, and $h(u)_L = h(u_L)$, whenever $u \in \widetilde{F}_{in\Sigma}(X, \Xi_\Sigma)^{(1)}$ and $u_L \in \Sigma_0$,

$h(u) \in \widetilde{F}_{out\Omega}(Y,\Xi_\Omega)^{(1)}$, and $h(u)_L = h(u_L)$, whenever $u \in \widetilde{F}_{out\Sigma}(X,\Xi_\Sigma)^{(1)}$ and $u_L \in \Sigma_0$,

$h(u) \in \widetilde{F}_{in\Omega}(Y,\Xi_\Omega)^{(1)} \cup \widetilde{F}_{in\Omega}(Y,\Xi_\Omega)^{(2)}$, whenever $u \in \widetilde{F}_{in\Sigma}(X,\Xi_\Sigma)^{(1)}$ and $u_L \in X$,

$h(u) \in \widetilde{F}_{out\Omega}(Y,\Xi_\Omega)^{(1)} \cup \widetilde{F}_{out\Omega}(Y,\Xi_\Omega)^{(2)}$, whenever $u \in \widetilde{F}_{out\Sigma}(X,\Xi_\Sigma)^{(1)}$ and $u_L \in X$,

$h(u) \in \widetilde{F}_{in\Omega}(Y,\Xi_\Omega)^{(2)}$, and $h(u)_L = h(u_L)$, whenever $u \in \widetilde{F}_{in\Sigma}(X,\Xi_\Sigma)^{(2)}$, and

$h(u) \in \widetilde{F}_{out\Omega}(Y,\Xi_\Omega)^{(2)}$, and $h(u)_L = h(u_L)$, whenever $u \in \widetilde{F}_{out\Sigma}(X,\Xi_\Sigma)^{(2)}$.

Relation h is said to be *down linear*, if the number of the positions of each letter of $E_{in}$ in $h_\Sigma(\sigma)$ is one at most whenever $\sigma \in \Sigma$; *up linear* is defined respectively for the letters in $E_{out}$. Relation h is *down preserving* (otherwise *down deleting*), if $|\mathcal{G}_{inh_\Sigma(\sigma)}| = |\mathcal{G}_\sigma|$ for each $\sigma \in \Sigma$, respectively is defined *up preserving and up deleting*. We call h *down alphabetic*, if $h(X \cup \Xi_\Sigma) \subseteq Y \cup \Xi_\Omega$, and for each $\sigma \in \Sigma$, $h_\Sigma(\sigma) = \omega(\varepsilon_i ; \varepsilon_j \mid i \in \mathcal{G}_{in}, j \in \mathcal{G}_{out})$, where $\omega \in \Omega$, cf. *tree homomorphism* Denecke K, Wismat SL (2002). Notice that because net homomorphism is in its nature "replacing", it can be seen as a special type of rewriting systems.

**Definition 1.3.2.** SUBSTITUTION.

Let T and S be arbitrary jungles and P a family of sets of positions. We define

$T(P \leftarrow S : *) = \bigcup (v(\mu_i^{(1)} v_i s; \lambda_j^{(1)} \delta_j s) : t = v(\mu_i; \lambda_j \mid i \in \mathcal{G}_v^{UN}, j \in \mathcal{G}_v^{UN}), p(t, \mu_{iL}) \in P,$

$p(t, \lambda_{jL}) \in P, t \in T, s \in S, *, v_i s \in s_{outg}, \delta_j s \in s_{ing})$.

That is $T(P \leftarrow S : *)$ is the jungle which is obtained by "replacing" (considering conditions *) all the subnets of each net t in T, having the position set in family P, by each net in S. Notice that the result may be T (that is no execution in replacing), if the arities of the replacing nets and on the other hand the arities of the nets in T are different or P does not represent any positions of subnets in nets of T.

If each position set of family V of subnets of each net t in T is wished to be replaced by each of elements in S, we write simply $T(V \leftarrow S)$.

Next introduced substitution relation is a special example of net homomorphisms, an essential component in rewriting.

*Net substitution relation* (here f) is such a net homomorphism in $F_\Sigma(X,\Xi) \cup \widetilde{F}_\Sigma(X,\Xi)$ that each ranked letter rewriting relation is identity relation, as well as the initial arity rewriting relations, and for each $v \in \widetilde{F}_\Sigma(X,\Xi)^{(2)})$ $f(v) = v^{(1)} v^{(2)} f(v_L)$ and for each $\mu \in \widetilde{F}_\Sigma(X,\Xi)^{(1)})$ $f(\mu) = \mu^{(1)} f(\mu_L)$.

Let $X_o$ be a set of frontier letters. Net substitution relation f is said to be $X_o$-*joining*, if

(i) $\{f(x)_L : x \in X_o\}$ is singleton and

(ii) the arity of the gluing form of each letter in $X_o$ is occupying an unoccupied arity of $f(x)_L$, and $|X_o|$ is the cardinality of the set of those unoccupied arities.

It is worth to remember that if an image of net substitution relation for a leaf is empty (set), the arity having been occupied by that leaf is after substitution an unoccupied arity.

Let P and T be arbitrary jungles. If S is a catenation of substitutions such that $T = S(P)$, we say that there is an S-*substitution route* between P and T.

**Definition 1.3.3.** INSTANCE. Net t is an *instance* of net s, if $t = f(s)$ for some net substitution relation f. Notice that s is a context of t for the in-glue form of net $f(v)_L$, whenever $v \in X \cup \Sigma_0$ is in s. Notice that $s = \text{con}_P(f(s))$, if $P \in \{ p(f(s),s), \cup(p(f(s),f(x)_L) : x \in X \cap L(s))\}$.

**Definition 1.3.4.** MATCHING. Net s is said to *match* t by net substitution relation f in $p(t,s)$, in a so called *matching point*, if $f(s) \in \text{sub}(t)$; thus $\text{apex}(s) \in \text{enc}(t)$. If net s matches net t, we say that the genuine tie-arities of s in the linkages between s and t are the *matching arities* of s in t, denoted MA(t,s).

## 1.4. §       Covers and Partitions

**Definition 1.4.1.** For jungle T a type $\rho$ of net (e.g. a tree) being in enc(T) is of *maximal* $\rho$-type in enc(T), if it is not an enclosement of any other $\rho$-type net in enc(T) than of itself. The other $\rho$-type nets in enc(T) are *genuine*.

**Definition 1.4.2.** COVER of NET. A set of nets is said to be a *cover* of net t, if each node of t is in a net of the set. We denote the set of all covers of net t with Cov(t), and for jungle, say T, we agree Cov(T) = ∪(Cov(t):t∈T).

**Definition 1.4.3.** SATURATION of NET. Cover A *saturates* net t, if A⊆enc(t). We denote the set of all saturating covers of net t with Sat(t), and for jungle, say T, we agree
Sat(T) = ∪(Sat(t):t∈T).

E.g. A saturating cover of net t is *natural*, if each net in the cover is of maximal ρ-type, where ρ-type net is the net the nodes having only one out-tie (resembling in that respect tree).

**Definition 1.4.4.** PARTITION of NET. A saturating cover of net t is a *partition* of t, if each node of t is exactly in one net in the cover. We reserve notation Par(t) as for the set of all partitions of net t, and for jungle, say T, we agree of notation Par(T) = ∪(Par(t):t∈T).

For an arbitrary jungle A we define *the partition induced by jungle* A

(denoted PI(A)) = { ⋒A´ ⋌ {⋒A´´: A´⊂A´´, A´´∈P(A)} : A´∈P(A)}.

We can write the following proposition:

**Proposition 1.4.** "A correlation between partitions and covers of nets".
For any net s and jungle E

  E∈Cov(s), if and only if  PI(E)⋒s∈Par(s).

Notice that if A is a saturating cover of net t, then PI(A) is a partition of t.

## 1.5. § REWRITE

In this chapter we introduce rewriting by using algebraic presentation described earlier regarding edges as ties or linkages which unite node- and on the other hand edge-rewriting (Thomas W (1997); Engelfriet J (1997)). *Term rewriting* as a special case of here presented rewriting can be probed e.g. in (Ohlebusch E (2002); Meseguer J, Goguen JA (1985)).

### 1.5.1

**Definition 1.5.1.1.** RULES. A *rewrite rule* is a set (possibly conditional) of ordered ´jungle-jungle´ -pairs (S,T), the elements of which are entitled *rule preforms,* simply *rules*, if there is no danger of confusion, denoted often by S→T, S is called the *left side* of pair (S,T) and T is the *right side* of it. We agree that right(R) is the set of all right sides of rule preforms in each element of set R of rewrite rules; left(R) is defined accordingly to right(R). The frontier letters of nets in those rule preforms are called *manoeuvre letters*.

### Types of rewrite rules

Next we shortly represent some general types of rewrite rules.

**Definition 1.5.1.2.** A rewrite rule is said to be *simultaneous*, if it is not a singleton van Glabbeek RJ (2001). The *inverse rule* of rule $\varphi$, $\varphi^{-1}$, is the set $\{(T,S) : (S,T) \in \varphi\}$. A rule is *single*, if it is singleton.

A rule is an *identity rule*, if the left side is the same as the right side in each rule preform of the rule. A rule is called *monadic*, if there is a net homomorphism connecting the left side to the right side in each rule preform of the rule. If for each rule preform $r$ in rule $\varphi$, $\mathrm{hg}(\mathrm{left}(r)) > \mathrm{hg}(\mathrm{right}(r))$, we call $\varphi$ *height_diminishing*, and if $\mathrm{hg}(\mathrm{left}(r)) < \mathrm{hg}(\mathrm{right}(r))$, $\varphi$ is *height increasing*; if $\mathrm{hg}(\mathrm{left}(r)) = \mathrm{hg}(\mathrm{right}(r))$, we call $\varphi$ *height saving*.

A rule is *alphabetically diminishing*, if for each rule preform $r$ in the rule there is in force:

(i) right($r$) is a ranked net or hg(right($r$)) = 0 or (ii) hg(left($r$)) = 2, root(right($r$)) ∈ L(left($r$)) and hg(right($r$)) = 1. For an abbreviation reason for a set of rules $\mathscr{R}$ we may use notation left($\mathscr{R}$) = { left($r$) : $r \in \varphi, \varphi \in \mathscr{R}$ } and respectively for the case of right side.

More specifically:

Definition 1.5.1.3 Any rule and the concerning pairs (i.e. rule preform) in it are said to be

1° *manoeuvre increasing*, if for each of its pairs, $r$, fron(left($r$)) ⊂ fron(right($r$)),

2° *manoeuvre deleting*, if for each of its pairs, $r$, fron(left($r$)) ⊃ fron(right($r$)),

3° *manoeuvre saving*, if for each of its pairs, $r$, fron(left($r$)) = fron(right($r$)),

4° *manoeuvre changing*, if at least for one of its pairs, $r$,

   fron(left($r$)) ⊈ fron(right($r$)) and fron(right($r$)) ⊈ fron(left($r$)),

5° *manoeuvre mightiness saving*, if for each of its pairs, $r$,

   |$p$(left($r$),x)| = |$p$(right($r$),x)|, whenever x is a manoeuvre letter,

6° *arity increasing*, if for each of its pairs, $r$, Uno(left($r$)) ⊂ Uno(right($r$)),

7° *arity deleting*, if for each of its pairs, $r$, Uno(left($r$)) ⊃ Uno(right($r$)),

8° *arity saving*, if for each of its pairs, $r$, Uno(left($r$)) = Uno(right($r$)),

9° *arity mightiness saving*, if for each of its pairs, $r$,

   |$p$(left($r$),ξ)| = |$p$(right($r$),ξ)|, whenever ξ is an unoccupied arity letter,

10° (ranked) *letter increasing*, if for each of its pairs, $r$, L(apex(left($r$))) ⊂ L(apex(right($r$))),

11° (ranked) *letter deleting*, if for each of its pairs, $r$, L(apex(left($r$))) ⊃ L(apex(right($r$))),

12° (ranked) *letter saving*, if for each of its pairs, $r$, L(apex(left($r$))) = L(apex(right($r$))),

13° (non-arity) *letter mightiness increasing*, if for at least one of its pairs, $r$,

   | ∪($p$(apex(left($r$)),z) : z is a frontier or ranked letter) | <

   | ∪($p$(apex(right($r$)),z) : z is a frontier or ranked letter) |,

14° X-*manoeuvre letter increasing, decreasing, saving*, if

   L(left($r$)) ∩ X   ⊂ , ⊃ , =  L(right($r$)) ∩ X , respectively,

15° X-*manoeuvre mightiness increasing, decreasing, saving*, if for each x∈X

|*p*(left(*r*),x)|  < , > , =  |*p*(right(*r*),x)|  , respectively.

Rule *φ* is *left linear*, if for each *r* ∈ *φ* and manoeuvre letter x there is in force |*p*(left(*r*),x)| = 1, and *right linear*, if  |*p*(right(*r*),x)| = 1. A rule is *totally linear*, if it is both left and right linear.

## 1.5.2 Renetting systems and Application

We introduce systems using rewrite rules to transform nets, and type wise to define sets of rules with special instructions regarding to apply them.

A set consisting of rewrite rules and of *conditional demands* (possibly none) Ohlebusch E (2002), (for the set of which we reserve symbol $\mathcal{C}$ ) to apply those rules is called *renetting system*, RNS in short, Engelfriet J (1997), and a ΣX-RNS, if its rewrite rules consist exclusively of pairs of ΣX-nets. Conditional demands may concern application orders cf. *process algebra with timing* Baeten JCM, Middelburg CA (2001), *probabilistic processes* Jonsson B, Yi W, Larsen KG (2001), *priority in process algebra* Cleaveland R, Lüttgen G, Natarajan V (2001). The objects to be applied may be required to possess certain nodes, linkages or neighbours or to be carrier nets for operations in selected algebras. Desired substitutions may be "context sensitive" i.e. chosen to be of left or right side and matching positions where applications are expected to be seen to happen may also be prerequisites. Notice that rules in RNS´s can be presented also exclusively by net types: pairs of rules in RNS´s defined in accordance with the amount of the arities or nodes possessed by them Engelfriet J (1997), *edge-replacing* Burkart O, Caucal D, Moller F, Steffen B (2001).

**Definition 1.5.2.1.** A *renetting system*, shortly entitled RNS, is *finite*, if the number of rules and $\mathcal{C}$ in it is finite. A RNS is said to be *limited*, if each rule of it is finite and in each pair of each rule the right side is finite and the heights of the nets in the both sides are finite. For the clarification we may use notation $\mathcal{C}(\mathcal{R})$ instead of $\mathcal{C}$ for RNS $\mathcal{R}$. A RNS is *conditional* (or *sensitive*), contradicted *nonconditional* or *free*, if its $\mathcal{C}$ is not empty. A RNS is *simultaneous*, contradicted *nonsimultaneous*, if it has a simultaneous rule.

A RNS is *elementary*, if for each pair r in each rule of the RNS is monadic or alphabetically diminishing. If each of the rules in a RNS is of the same type, the RNS is said to be of that type, too. For each RNS $\mathcal{R}$ we denote $\mathcal{R}^{-1} = \mathcal{R}(\varphi \leftarrow \varphi^{-1})$.

**Definition 1.5.2.2.**  APPLICATION TYPES. For given RNS $\mathcal{R}$, jungle S is $\mathcal{R}$-*rewritten* to jungle T (*rewrite result*), denoted S $\rightarrow_{\mathcal{R}}$ T (called $\mathcal{R}$–*application*), and is *reduced under* $\mathcal{R}$ or by rule $\varphi$ in $\mathcal{R}$, and is said to be a *rewrite object*  for $\mathcal{R}$ or $\varphi$ respectively, denoting   T = S$\varphi$ (the postfix notation is prerequisite), if the following "rewrite" is fulfilled:

T = $\cup$(S($\wp \leftarrow$ (right(*r*))g ) : left(*r*)  matches s in $\wp$ by some net substitution mapping $f_{s\wp}$, $r \in \varphi$, g $\in G_{s\wp}$ , $\wp \in p(s)$, s $\in$ S, $\mathcal{C}(\mathcal{R})$), where $G_{s\wp}$´s are sets of net substitution relations. Mapping $f_{s\wp}$ is called *left side substitution relation* and each g in $G_{s\wp}$ is *right side substitution relation*, c.f. under conditional demands "extra variables on right-hand sides" *conditional Rewrite Systems* Ohlebusch E (2002). We say that RNS is S-*instance sensitive* (S-INRNS), if for a rule $\varphi \in$RNS and for each s$\in$S, $\wp \in p(s)$, $G_{s\wp} \neq f_{s\wp}$ , and S-*mapping instance sensitive* (S-MINRNS), if right side substitution relations are mappings. If furthermore all right side substitution mappings are singletons, we entitle SingMINRNS to indicate RNS´s of that nature. If all rules in RNS are obligated to satisfy the demands, *instance sensitiveness* of RNS is said to be *thorough*. Notice that for substitution relations, $\mathcal{C}(\mathcal{R})$ may contain some orders liable to substituting manoeuvre letters in the rewrite process (*substitution order*), especially if rewrite objects have outside loops with the apexes of left sides of pairs in rules or $\mathcal{R}$ is manoeuvre increasing and instance sensitive. Instructions concerning binding right side substitution relations to specific rules in RNS might also have been included in $\mathcal{C}(\mathcal{R})$.

We say that $\mathcal{R}$ *matches a rewrite object*, if the left side of a rule preform matches it. We say that S is a *root* of T *in* $\mathcal{R}$ and T is a *result* of S *in* $\mathcal{R}$. Observe that T = S, if $\mathcal{R}$ does not match S; of course $\mathcal{C}$ may contain demands for necessary matching. The enclosements at which rewrites can take places (the sets of the apexes of the left sides in the pairs of the rules in RNS´s) satisfying all requirements set on the RNS are called the *redexes* of the concerning rules or RNS´s in the rewrite objects. For RNS $\mathcal{R}$ and jungle S we denote

S$\mathcal{R}$ =  $\cup$(S$\varphi$ : $\varphi \in \mathcal{R}$ ).

Rule $\varphi$ of $\mathcal{R}$ is said to be *applied* to jungle S, if for each s∈S, s has *φ-redexes* (redexes of $\varphi$ in s) fulfilling $\mathcal{C}(\mathcal{R})$ and thus $\varphi$ is *applicable to* S and S is *φ-applicable* . RNS $\mathcal{R}$ is *applicable to* S and S is *$\mathcal{R}$-applicable* , if $\mathcal{R}$ contains a rule applicable to jungle S.

For RNS $\mathcal{R}$ we define $\mathcal{R}$−*transformation relation* on $F_\Sigma(X,\Xi)$

$$\to_\mathcal{R} = \{(s,s\mathcal{R}) : s\in F_\Sigma(X,\Xi)\}.$$

**Lemma 1.5.2.** Any relation can be presented with a RNS and its rewrite objects. On the other hand with any given RNS we have RNS-transformation relation.

PROOF. Let r be a relation. Constructing RNS $\mathcal{R} = \{a \to b : (a,b)\in r\}$ we obtain

$$r = \{(a,a(a \to b)): a \to b \in \mathcal{R} \}. \square$$

It is quite clear that a net cannot be $\mathcal{R}$-rewritten, if $\mathcal{R}$ is not instance sensitive and the matching points of the left sides of the pairs in the rules of $\mathcal{R}$ have outside loops to the net, because the apexes of the right sides of those pairs must be enclosed in some images of the right side substitution relations.

**Definition 1.5.2.3.** We call RNS *feedbacking in respect to a net*, if while applying a rule in it for that net, elements in the image sets of each right side substitution relation regarding to the preforms in the involving rule overlap that net; feedbacking for a rule is *total*, if the demands concern all elements in the image sets (always total, if the substitution relations are mappings since the image sets are then singletons) and *partial*, if RNS is feedbacking but not totally. If instead of only overlapping, we claim the enclosement condition for elements in the sets of the right side substitution images, feedbacking RNS is *innerly feedbacking* - which is e.g. the case in not instance sensitive RNS´s – and if no overlapping is enclosement, RNS is *outherly feedbacking*. If the net in concern of feedbacking is the applicant for RNS, we speak of *self feedbacking*. The form of innerly self feedbacking RNS in respect to a net, say t, where for each rule preform *r* there is in force equation t*r* ⌞ apex(right(*r*)) = t ⌞ apex(left(*r*)), we name *environmentally saving* in respect to the rewrite object in concern. If all rules in RNS satisfy the feedbacking demands we speak of *thoroughly feedbacking* RNS. It is worth to remind that INRNS´s are capable to join distinct applicable nets.

By the cardinality of the image sets of right side substitution relations and on the other hand for each left side substitution mapping by the cardinality of the set of right side substitution relations regarding to mutual rules, the necessary rules in RNS´s can be compensated so not to exceed finite number – the right side substitutions relations can be defined type wise, i.e. setting their image sets to consist nets of certain type (e.g. limited number of nodes or unoccupied arities). In feedbacking RNS´s right side substitution relations may be regulated to depend on the type of rewrite objects (thus covering large portions of object nets by a limited number of regulations and not needing to raise the amount of rules possibly to infinite), e.g. replacing manoeuvre letters, existing only in the right-handed sides of pairs in rules, by overlapping nets in specific positions, if any.

It is somewhat of worth to mention that RNS´s, not instance sensitive, can own the same rewriting power than INRNS´s, but then we may be compelled to accept infinite number of manoeuvre altering rules – e.g. in the case we have a manoeuvre letter increasing INRNS, where for left side substitution mapping f and right side substitution relation g, $g(y)_L$ overlaps $f(x)_L$ for some manoeuvre letters $x \neq y$ (i.e. rewrite results are expected to contain loops) and there is expected to be an infinite number of rewrite objects for which RNS is to be constructed, or if the cardinality of set $\{f(x)_L : x \in X\}$ is infinite.

The following example offers a manifestation of particularity in substitution orders:
Let $\xi_{a1}, \xi_{b1}, \xi_{c1}, \xi_{d1}$ be out-arities and $\xi_{a2}, \xi_{b2}, \xi_{c2}, \xi_{d2}$ are in-arities, f standing for a left side and g a right side substitution relation,

$f(x)_L, g(x)_L \in [S], f(x) = g(x), S = d(\xi_{d2}\xi_{c1}c(\xi_{c2}; \xi_{c1}); \xi_{d1})$

$f(x) = \xi_{c2}s_1$, $s_1 = c(\xi_{c2}; \xi_{c1}\xi_{d2} d(\xi_{d2}; \xi_{d1}))$ $(\in [S])$, $s_1$ is a representative of S,

$g(y) = \xi_{d1}d(\xi_{d2}\xi_{c1}t_1; \xi_{d1})$, $t_1 = c(\xi_{c2}\xi_{b1}t_2; \xi_{c1}\xi_{d2}S)$, $t_2 = b(\xi_{b2}g(y); \xi_{b1})$,

and $g(y)_L$ and $f(x)_L$ are overlapping each other, if possible,

$r = a(\xi_{a2}; \xi_{a1}x) \rightarrow b(\xi_{b2}y; \xi_{b1}x)$.

If x is substituted first, the result offers fixing point for y-substitution, yielding a loop structure as a result. If on the other hand y is firstly substituted, the result is totally of a different nature, where there is a continuously growing chain of iterated nets via y-substitutions.

For left side substitution mapping f in loop situations between images must be $f(x)_L$ overlapping $f(y)_L$ for some manoeuvre letters $x \neq y$, and one of them must contain itself as a subnet; illustrated in the next example of an application of manoeuvre cardinality increasing, not instance sensitive rule, with rewrite object containing a loop.

Let $\xi_{a1}, \xi_{b1}, \xi_{\alpha1}$ be out-arities and $\xi_{a2}, \xi_{b2}, \xi_{\alpha2}$ are in-arities,

$f(x) = \xi_{a2} t_1$, $t_1 = a(\xi_{a2}; \xi_{a1}\xi_{b2} t_2)$, $t_2 = b(\xi_{b2}; \xi_{b1}\xi_{\alpha2} t_3)$, $t_3 = \alpha(\xi_{\alpha2}; \xi_{\alpha1} f(x))$

$f(y) = \xi_{b1} b(\xi_{b2}\xi_{a1} a(\xi_{a2}; \xi_{a1}) ; \xi_{b1})$

and $f(x)_L$ and $f(y)_L$ are overlapping each other. The result via rule $\alpha(x;y) \to \beta(x;y,y)$ is unaffected by the substitution orders between x and y.

In the following our presumption for RNS´s are not to be instance sensitive, if not indicated otherwise.

**Definition 1.5.2.4.** DERIVATION. *Derivation in set $\mathcal{R}$ of RNS´s* is any catenation of applications of RNS´s in $\mathcal{R}$ , say $\mathcal{D}$, such that the rewrite result of the former part is the rewrite object of the latter part of the consecutive elements in the catenation. The rewrite results of the elements in the catenation are called $\mathcal{D}$-*derivatives* of the rewrite object for the first element, and the catenation of the corresponding rules is entitled *deriving sequence* in $\mathcal{R}$ , for which in an operational use the postfix notation is the default. We agree of the associativity that for any deriving sequence $q$ and any jungle S

$$Sq = (Sq_1)q_2 , \text{ if } q = q_1 q_2 .$$

## 1.5.3  Transducers and the Types

**Definition 1.5.3.1.** TRANSDUCER. For each $\omega \in \Omega$, $i \in \mathcal{I}_\omega$ and $j \in \mathcal{J}_\omega$ , let r be a bijection, *RNS-attaching mapping*, joining a set of RNS´s to each triple $(\omega, i, j)$. Let $\mathcal{A} = (F_\Sigma(X,\Xi), \Omega_\mathcal{A} Y_\mathcal{A} \Xi_\mathcal{A})$ be a $\Omega_\mathcal{A} Y_\mathcal{A} \Xi_\mathcal{A}$-algebra, where for each $\omega \in \Omega$

$\omega^\mathcal{A}: F_\Sigma(X,\Xi)^\alpha \otimes \widetilde{F}_{out\Sigma}(X,\Xi)^\beta \mapsto F_\Sigma(X,\Xi)$, where $\alpha = $ in-rank$(\omega)$ and $\beta = $ out-rank$(\omega)$,

is such an operation relation that

$\omega^{\mathcal{A}}(s_{iL}; \lambda_j \mid i \in \mathcal{I}_\omega, j \in \mathcal{J}_\omega) \subseteq \bigcup (s_{iL} r(\omega, i, j): i \in \mathcal{I}_\omega, j \in \mathcal{J}_\omega)$.

$\mathcal{A}$ is called *a renetting algebra*. For any net $t \in F_\Omega(X,\Xi)$ realization $t^{\mathcal{A}}$ is called R-*transducer* (R-TD) over *RNS-attached family* $R = \{r(\omega,i,j) : \omega \in \Omega \cap L(t), i \in \mathcal{I}_\omega, j \in \mathcal{J}_\omega\}$ of sets of RNS´s and it is also entitled an *interaction* between those RNS´s. Notice that $\omega^{\mathcal{A}}$ with in-rank($\omega$) = out-rank($\omega$) = 1, represents RNS-transformation relation. Referring to a set of TD´s, say G, in concern for the realizations, we use notation $(F_\Sigma(X,\Xi),G)$ for the renetting algebra. We want to notify that samples of possible conditions liable to realizations of upstream subnets in carrier nets of transducers may be used to set extra demands for selecting desired operating RNS´s to influence data flows from targeted in-arities. That notion is expressed in the next lemma.

We say that a TD *matches a rewrite object*, if any of its RNS does it. Let $\mathcal{I}$ be an arbitrary index set, and for each $i \in \mathcal{I}$ let $\mathcal{R}_i$ be a TD, thus we denote Cartesian element $\overline{\mathcal{R}}(\mathcal{I}) = (\mathcal{R}_i: i \in \mathcal{I})$, and $\overline{a}\overline{\mathcal{R}}(\mathcal{I}) = (e[\mathcal{I}](i,\overline{a})\mathcal{R}_i : i \in \mathcal{I})$, whenever $\overline{a}$ is a Cartesian element. For any applicant S  $S\mathcal{R}$  is called the *result of* S *in* $\mathcal{R}$.

**Lemma 1.5.3.1.** The conditional demands for TD´s can be presented as a TD´s having no demands, and thus any TD, let us say $\mathcal{R}$, can be given as a TD with no demands and the carrier net of that TD having the enclosements of the carrier net of $\mathcal{R}$ in its enclosements.

PROOF. The claim is following from lemmas 1.2.3 and 1.5.2. □

**Definition 1.5.3.2.** TRANSDUCER TYPES. If each RNS in a TD is of the same type (e.g. manoeuvre saving), we say that the TD is of that type. A TD is said to be *altering*, if while applying it is changing, e.g. the number of the rules in its RNS´s is changing (thus being *rule number altering*). A TD is entitled *contents expanding*, if some of its RNS´s contain a letter mightiness increasing rule preform. A TD is called *trivial*, if each rewrite objects for it is the same as the result in the TD. A TD is called *upside down tree TD*, if each ranked letter in the carrying net of the TD has only one in-arity.

A TD is a *transducer graph* (TDG) over a set of transducers, if the set of the carrying nets of all transducers in the set is a partition of the carrying net of the TD. I.e. TDG is a special case among transformer graphs. Any transducer graph over set T is denoted TDG(T), and any TDG(T) is *beyond each subset of* T, analogous with TG relevant to that issue.

A TDG(T) is entitled *direct* (in contradiction to *indirect* in other cases), if the only claims for the TDG(T) are those of the TD´s in T.

Any TDG over a set can be visualized as a TG over the same set.

**Lemma 1.5.3.2.** The carrying net of any altering TD can be seen as an enclosure of the larger carrying net of some nonaltering TD.

PROOF.  Straightforwardly from lemma 1.5.3.1.  □

**Definition 1.5.3.4.** TD-TRANSFORMATION RELATION. Let $\mathcal{R}$ be a transducer. We define $\mathcal{R}$-*transformation relation* $\to_{\mathcal{R}}$ in the set of the jungles such that

$$\to_{\mathcal{R}} = \{(t, t\mathcal{R}) : t \text{ is a jungle}\}.$$

We say that two transducer $\mathcal{P}$ and $\mathcal{R}$ are the same, $\mathcal{P} = \mathcal{R}$, if $\to_{\mathcal{P}} = \to_{\mathcal{R}}$.

**Definition 1.5.3.3.** NORMAL FORM and CATENATION CLOSURE. $\mathcal{D}(\mathcal{R})$ is the notation for the set of all derivations in TD $\mathcal{R}$. If for jungle S and TD $\mathcal{R}$, $S\mathcal{R} = S$, S is entitled $\mathcal{R}$-*irreducible* or of *normal form under* $\mathcal{R}$. For the set of all $\mathcal{R}$-irreducible nets we reserve the notation IRR($\mathcal{R}$). For each jungle S and TD $\mathcal{R}$ we denote the following:

$$S\mathcal{R}^{\wedge} = S\mathcal{R}^{*} \cap \text{IRR}(\mathcal{R}),$$

where $\mathcal{R}^{*}$, *the catenation closure of* $\mathcal{R}$, is the transitive closure of the rules in $\mathcal{R}$.

Let $\mathcal{R}$ be a TD over family $\mathscr{R}$. We define *normal form TD of* $\mathcal{R}$, TD$^{\wedge}$,

$$\mathcal{R}^{\wedge} = \mathcal{R}(\mathscr{R} \leftarrow \mathscr{R}^{\wedge} : \mathscr{R} \in \mathscr{R}).$$

# 1.6. § Equations and decompositions as examples of TD´s

**Definition 1.6.1.** EXPLICIT and IMPLICIT RNS-CLAUSE. Let $\mathcal{R}$ and $\mathcal{G}$ be two TD´s. Let H be a list of symbols in $\otimes$, $\mathcal{R}$ and $\mathcal{G}$, where $\otimes = \{=, \in, \subset, \subseteq\}$. If $\mathcal{R} \otimes_e \mathcal{G}$, where $\otimes_e \in \otimes$, we call TDG over $\mathcal{R}, \mathcal{G}, \otimes_e$ a *RNS-clause* (RCl), denoted $\mathcal{E}(\mathcal{R}, \mathcal{G}, \otimes_e)$. $\mathcal{E}(\mathcal{R}, \mathcal{G}, \otimes)$ is of *first order* in respect to an element of H, if that element exists only once in the equation.

RNS-clauses cover also the ´ordinary´ equations (with no RNS´s), being due to lemma 1.5.2.

Any TD in RNS-clause $\mathcal{E}(\mathcal{R}, \mathcal{G}, \otimes)$ is called a *factor*; a *left handed* factor or *a factor of $\mathcal{R}$*, if it exists exclusively in $\mathcal{R}$, and a *right handed* factor or *a factor of $\mathcal{G}$*, if it exists exclusively in $\mathcal{G}$.

Let K be a factor in RNS-clause $\mathcal{E}(\mathcal{R}, \mathcal{G}, \otimes)$. We say that the RCl is a *representation* of K; specifically an *explicit* one (in contradiction to *implicit* in other cases), if K=$\mathcal{R}$ and K is not a factor of $\mathcal{G}$. The right handed factors are *composers* of $\mathcal{G}$ is a *compositions* of K, if $\mathcal{E}(\mathcal{R}, \mathcal{G}, \otimes)$ is an explicit representation of K, and $\otimes$ is = . A composition of K is said to be *linear/nonlinear*, if it is a direct/an indirect TDG. Because each operation in nets can stands for a simple case of TD´s then that simplyfied RNS-clause equates ordinary equations with operations of variables.

The question in automated problem solving basically is how to generate nets from enclosements of a probed net those enclosements being in such a relation with the enclosements in the conceptual nets that the particular relation is invariant under that generating transformation i.e. preserves invariability under class-rewriting. Therefore in the next three chapters we handle an idea of automated problem solving, as formal inventiveness. In problem solving, an essential thing is to see over details, and that is the task we next grip ourselves into by describing ideas such as partitioning nets by RNS´s and a connection between partitions by introducing the abstraction relation. We concentrate to construct TD-models for formulas of jungle pairs by conceptualizing ground subjects and then reversing counterparts of existing TD-solutions back to ground level. Then we widen the solution hunting by classifying intervening TDG-derivations. Finally we formulate abstract quotient algebra based on congruence class rewriting operations.

## 2. § <u>Inventiveness</u>

## 2.1. Recognizers and Languages

**Definition 2.1.1** RECOGNIZER and RECOGNITION. Let A and B be sets and let α: A ↦ B be a binary relation and A´ a subset of B. We define *recognizer* $\mathcal{A}$ such that $\mathcal{A} = (α, A´)$ entitling α a *recognizer relation* and A´ a *final set*. Element of A, s (*probed object*), is said to be *recognized* by recognizer $\mathcal{A}$, if sα∈A´. *Language* $\mathcal{L}_\mathcal{A}$ is the set of the elements recognized by $\mathcal{A}$, i.e. α separates from probed objects those ones, which have property A´. As a special case for nets the recognizer relation can chosen to be χη, where χ is a net homomorphism and η a relation transforming nets to wanted realizations of them, cf. *tree automata rules or tree recognizer homomorphisms* Gécseg F, Steinby M (1997), *hyper tree recognizer hypersubstitutions* Denecke K, Wismat SL (2002).

In general: Set *H satisfies transformer $\mathcal{T}$ via recognizer $\mathcal{A}$* or is a $\mathcal{A}$-*model of formula $\mathcal{T}$*, denoted H $\models^\mathcal{A}$ $\mathcal{T}$, if $\mathcal{A}$ recognizers $\mathcal{T}$–transformation of H, $\mathcal{T}$(H). E.g. a recognizer relation (automorphism) in $\mathcal{A}$ giving desired truth value from $\mathcal{T}$(H) we can say that H is $\mathcal{A}$-*solution for $\mathcal{T}$*, if the value given by the recognizer mapping is true. I.e. "validity of Boolean inference": the nest of transformer consisting of elementary logical relations (Boolean) and the concerning recognizer relation being "truth values giving automorphism from truth values of variables in the carrier net of the transformer", the final set consequently represents the value "true" or "untrue". A transformer can also be RCl and H a factor in it Chang CC, Keisler HJ (1973).

For nets S, T and TD $\mathcal{R}$ in model theoretical notation $\mathcal{R} \models^\mathcal{A}$ (S,T) $\mathcal{R}$ is named a $\mathcal{A}$-*model of formula* (S,T), or pair (S,T) is a $\mathcal{A}$-model of formula $\mathcal{R}$ (denoted in that interpretation (S,T) $\models^\mathcal{A} \mathcal{R}$ ), if recognizer $\mathcal{A}$ (e.g. probing the truth values) is recognizing RCl T⊆S$\mathcal{R}$. Cf. inferring *winning game graphs* Thomas W (1997).

**Definition 2.1.2.** Here we introduce a convenient tool, needed later in the context of abstraction relation. Let $\mathcal{I}$ be an arbitrary index set and for each $i,j \in \mathcal{I}$ let $\theta_{ij}: A_i \mapsto A_j$ be a binary relation from set $A_i$ to set $A_j$. Let $\overline{A} = \Pi(A_i: i \in \mathcal{I})$ and $\tilde{\theta} = \Pi(\theta_{ij} : (i,j) \in \mathcal{I}_\theta)$ for some $\mathcal{I}_\theta \subseteq \mathcal{I}^2$. Let $\alpha: \overline{A} \mapsto \Pi(\theta_{ij} : (i,j) \in \mathcal{I}^2)$ be a binary relation, where $\bar{a}\alpha = \Pi(\theta_{ij} : (i,j) \in \mathcal{I}^2, \mathrm{elem}(i,\bar{a})\, \theta_{ij}\, \mathrm{elem}(j,\bar{a}))$, whenever $\bar{a} \in \overline{A}$. The language recognized by $\mathcal{A} = (\alpha, \tilde{\theta})$ is $\tilde{\theta}$-*associated* over $\mathcal{I}_\theta$ (denoted $\mathcal{L}_{\tilde{\theta}}$); if in $\tilde{\theta}$ all $\theta_{ij}$ are the same, say $\theta$, we speak of $\theta$-*associated language*.

In other words this recognizer picks from among $\overline{A}$ such elements, the projection elements of which are pair wise in a relation of set $\{\theta_{ij} : (i,j) \in \mathcal{I}_\theta\}$. Notice that $\theta$-associated language over a singleton is $\theta$-relation itself, if $|\mathcal{I}| = 2$.

## 2.2 Problem and Solution

**Definition 2.2.1.** *Problem* $\mathfrak{I}$ is a triple $(S, \mathcal{A}, \mathcal{C})$, where the *subject of the problem* $S$ is a jungle, a set of *mother nets*, $\mathcal{A}$ is a recognizer and *limit demands* $\mathcal{C}$ ($\mathcal{C}(\mathfrak{I})$ precise notation, if necessary) is a sample of prerequisites to be satisfied in recognition processes. TD $\mathcal{V}(\mathfrak{I})$ is a *presolution* of problem $\mathfrak{I}$, if $S\mathcal{V}(\mathfrak{I}) \in \mathcal{L}_\mathcal{A}$, thus $S\mathcal{V}(\mathfrak{I})$ being called a *solution product*, and if furthermore $\mathcal{V}(\mathfrak{I})$ fulfils the demands in set $\mathcal{C}$, $\mathcal{V}(\mathfrak{I})$ is a *solution* of $\mathfrak{I}$. E.g. solution $\mathcal{V}$ may be a system, by which from certain circumstances $S$, with some limit demands (e.g. the number of the steps in the process) can be built surrounding $S\mathcal{V}$, which in certain state $\alpha(S\mathcal{V})$ (for morphism $\alpha$ of a recognizer) has a capacity characterized by the type of the elements in the final set of the recognizer.

We can describe a solution for a problem as wandering in a net:

    1. Starting from a given net node (mother net)

    2. to the acceptable net (solution product) ($\in \mathcal{L}_\mathcal{A}$) of the TFG

    3. via the right route in the RPG (TDG-solution) (limit demands accomplishments).

Cf. Aceto L, Fokkink WJ, Verhoef C (2001)   mother net $\vDash \langle TD \rangle \mathcal{A}$.

# 3.§ Parallel Process and Abstract Algebras

(for *Automated Problem Solving*)

## 3.1. Partition RNS and Abstraction Relation

**Definition 3.1.1.** PARTITION RNS. For each jungle (here c) we define a *partition RNS* (PRNS) $\mathcal{W}$ of that jungle as a RNS fulfilling conditions (i)-(iii):

(i) $\mathcal{W}$ is manoeuvre mightiness and arity mightiness saving, not instance sensitive,

(ii) L(apex(right($r$)))\Ξ is a singleton and its element is outside L(c), whenever $r \in \varphi$, $\varphi \in \mathcal{W}$, and {(left($r$),right($r$)): $r \in \varphi$, $\varphi \in \mathcal{W}$} is an injection,

(iii) $\mathcal{C}(\mathcal{W}) \supseteq \{L(c) \cap L(c\mathcal{W}\hat{\ }) = \varnothing\}$.

In a special case where the left sides of rule preforms do not overlap each other, {apex(left($r$)): $r \in \varphi$, $\varphi \in \mathcal{W}$} is a partition of jungle c. Furthermore be notified that a characteristically feature of PRNS´s is that $L(c\mathcal{W}\hat{\ })(\mathcal{W}^{-1})\hat{\ }$ is a partition of net c . We say that $c\mathcal{W}\hat{\ }$ is $\mathcal{W}$-*partition result* from c. Observe that for each PRNS there may be several jungles, the PRNS´s of which it is an example of, the nets of those jungles having apexes of left sides of rule preforms of that PRNS in different positions. One of the important factors regarding to the partition result is the independence of *reduction ordering* Jantzen M (1997), *partial matching* Körner E, Gewalting M-O, Körner U, Richter A, Rodemann T (1999).

The next characterization clause 3.1 says that the necessary and sufficient condition in order to be the partition result of a PRNS for a rewrite object is that there is a one-to-one correlation between the elements of the partition of that rewrite object and the letters of the result in respect to the cardinality of the positions of the unoccupied arities.

**Proposition 3.1.** "Characterization Clause". Let a and b be jungles. Then

(i) ( $\forall$ P∈Par(a) ) ( $\exists$ n ∈ { $\delta_D(\alpha)$ : $\alpha \in$ enc(b), L($\alpha$)\Ξ is a singleton } $\cup$ { $\delta_D(t)$ : t∈P } )

$|\bigcup( p(P,t) : \delta_D(t) = n, t \in P ) | \neq |\{c : |L(c) \cap \Xi| = n, c \in \text{enc}(b), L(c) \backslash \Xi \text{ is a singleton}\}|$,

or  (ii)  $(L(a)\setminus \Xi) \cap (L(b)\setminus \Xi) \neq \emptyset$,

if and only if

$a \mathscr{R}^{\hat{}} \neq b$, whenever $\mathscr{R}$ is a PRNS.

PROOF.   Our characterization (i) liable to net placing numbers is originated from PRNS definition item (iii), because of manoeuvre mightiness and arity mightiness saving feature of PRNS and characterization (ii) is a subject to definition item (ii). □

**Definition 3.1.2.** SUBSTANCE and CONCEPT.  If for jungles s and t and PRNS $\mathcal{W}$ of s there is equation $s\mathcal{W}^{\hat{}} = t$, we say that rewrite object s is a *substance* of t via $\mathcal{W}$, and rewrite result t is the *concept* of s via $\mathcal{W}$.

**Lemma 3.1.**  For each jungle c and each PRNS $\mathcal{W}$ of c

$$c\mathcal{W}^{\hat{}}(\mathcal{W}^{-1})^{\hat{}} = c$$

PROOF.   Straightforward due to non-deleting rules and (iii)-condition in PRNS´s yielding the partition result is independent of reduction orders. □

"The abstraction relation" to be presented next, is needed in the process to refer to a common origin via PRNS between the subjects in problems to be solved and jungles presenting known solutions.

**Definition 3.1.3.**   ABSTRACTION RELATION.  The *abstraction relation* (AR) is such a binary relation of the pairs of jungles, where for each pair (here (s,t)) there is such substance c and *intervening* PRNS $\mathcal{W}_1$ and $\mathcal{W}_2$, that

$c\mathcal{W}_1^{\hat{}} = s$    and    $c\mathcal{W}_2^{\hat{}} = t$.

Concepts  s and t are said to be *abstract sisters* with each other and c is entitled a *common origin* for s and t.

**Theorem 3.1.**  "A characterization of the abstraction relation". Let θ be the abstraction relation, and a and b be two jungles. Thus

$a \theta b \iff \delta_D(a) = \delta_D(b)$.

PROOF. ´⇐´:

The proof is executed in a finite case and for nets instead of jungles, but that does not diminish the power of the proof. Let $A_1$, $A_2$, $B_1$, $B_2$, and $B_3$ be such jungles that $A_1 \cup A_2$ is a partition of net a, and $B_1 \cup B_2 \cup B_3$ is a partition of net b. We can construct substance c for a and b as in the following figures, distinguished in two different cases depending on the positions of unoccupied arities.

For border $\mathscr{H}_{12}$ in the partition of net a and borders $\mathscr{B}_{12}$ and $\mathscr{B}_{23}$ in the partition of net b it is to be constructed net c and partitions for it, where

(i) A´-partition: $A_1´ \cup A_2´$, where $|A_1´| \geq |A_1|$, $|A_2´| \geq |A_2|$, and there is bijection

$f_a: A_1´ \cup A_2´ \mapsto A_1 \cup A_2$ such that $|L(a´)| \geq |L(f_a(a´))|$ whenever $a´ \in A_1´ \cup A_2´$, and

(ii) B´-partition: $B_1´ \cup B_2´ \cup B_3´$, where $|B_1´| \geq |B_1|$, $|B_2´| \geq |B_2|$ and $|B_3´| \geq |B_3|$, and there is bijection

$f_b: B_1´ \cup B_2´ \cup B_3´ \mapsto B_1 \cup B_2 \cup B_3$ such that $|L(b´)| \geq |L(f_b(b´))|$ whenever

$b´ \in B_1´ \cup B_2´ \cup B_3´$, and

(iii) border $\mathscr{H}_{12}´$ " a subset of the set of the linkages of the nets in $B_2´$ " and borders $\mathscr{B}_{12}´$ and $\mathscr{B}_{23}´$ " a subset of the set of the linkages of the nets in A´-partitions " fulfil the equations: $|\mathscr{H}_{12}´| = |\mathscr{H}_{12}|$, $|\mathscr{B}_{12}´| = |\mathscr{B}_{12}|$, $|\mathscr{B}_{23}´| = |\mathscr{B}_{23}|$, and

(iv) $\Lambda_1$, $\Lambda_1´$ and $\Lambda_2$, $\Lambda_2´$ are sets of unoccupied arities positioned as shown in cases 1° and 2°.

Straightforwardly we thus can construct PRNS $\mathcal{W}_a$ and $\mathcal{W}_b$ of net c such that

$A_1´\mathcal{W}_a\hat{} = A_1$, $A_2´\mathcal{W}_a\hat{} = A_2$, $B_1´\mathcal{W}_b\hat{} = B_1$, $B_2´\mathcal{W}_b\hat{} = B_2$ and $B_3´\mathcal{W}_b\hat{} = B_3$.

Case 1° The unoccupied arities are in neighbouring elements in a partition of net b.
Case 2° The unoccupied arities are in such elements of a partition of net b which are totally isolated from each other.

PROOF. ´⇒´:

Let us in contradiction suppose $\delta_D(a) \neq \delta_D(b)$. If c is a substance for net a, we have $\delta_D(c) = \delta_D(a)$, because the PRNS between a and c is arity mightiness saving, and from the same reason we are not able to get any concept to c with the cardinality of the unoccupied arities differing from that cardinality of c. Therefore $(a,b) \notin \theta$. □

**Corollary 3.1.1.** Any substance and any of its concepts are in the abstraction relation with each other.

PROOF. Any substance and its concepts have the same amount of unoccupied arities, because intervening PRNS´s are arity mightiness saving. □

**Corollary 3.1.2.** The abstraction relation is an equivalence relation.

PROOF. Theorem 3.1. □

## 3.2. Altering RNS

"Macros" treated in this chapter are needed in process to get solutions for elements in the subject of the problem in study via known solutions in memories for problems the subject consisting nets with other elements than in the original subject.

**Theorem** 3.2. "Altering macro RNS-theorem". Let $\mathscr{R}$ be a RNS, nonconditional for the sake of simplicity, let t be an arbitrary jungle and $\mathscr{W}$ a nonconditional PRNS of t. Then there is such RNS $\mathscr{R}_\mathscr{W}$ and such a PRNS $\mathscr{W}_o$ of $t\mathscr{R}\hat{\,}$ that there is in force an implicit equation of first order for unknown $\mathscr{R}_\mathscr{W}\hat{\,}$, where $\mathscr{R}_\mathscr{W}$ is a composer for a linear composition of $\mathscr{R}\hat{\,}$:

$$t\,\mathscr{W}\hat{\,}\,\mathscr{R}_\mathscr{W}\hat{\,}(\mathscr{W}_o^{-1})\hat{\,} = t\,\mathscr{R}\hat{\,}.$$

We can also solve unknown RNS $\mathscr{R}$ from the explicit equation above for $\mathscr{R}\hat{\,}$ with suitable PRNS $\mathscr{W}_o$ of $t\,\mathscr{R}\hat{\,}$, if PRNS $\mathscr{W}$ and RNS $\mathscr{R}_\mathscr{W}$ are given.

PROOF. Without loosing of generality we present the proof keeping nets as rewrite objects instead of jungles.

1° First we prove the implicit equation.

Let t be an arbitrary net and $\mathcal{W}$ be a nonconditional PRNS and $\mathcal{R}$ an arbitrary nonconditional RNS. Let $\mathcal{I}$ be such an injection that joins an index element to each rule preforms of any rule, such that $\varphi = \{a_i \rightarrow B_i : i \in \mathcal{I}(\varphi)\}$, whenever $\varphi \in \mathcal{R}$. Let us next construct required $\mathcal{R}_\mathcal{W}$, a rule number altering *macro RNS* for $\mathcal{R}$ in regard to $\mathcal{W}$, (thus $\mathcal{R}$ being entitled as one of the *micro RNS´s* of $\mathcal{R}_\mathcal{W}$).

We denote jungle $G_o(t,\mathcal{W}) = (L(t\mathcal{W}\hat{\ }))(\mathcal{W}^{-1})\hat{\ }$. Hereby on the basis of lemma 3.1.3 we achieve $G_o(t,\mathcal{W}) \cap s \in Par(s)$, whenever $s \in enc(t)$. Let us consider rule preform $a_i \rightarrow B_i$, $i \in \mathcal{I}(\varphi)$, $\varphi \in \mathcal{R}$. Let $c_i$ be the context of a representative in $[t]$ for $apex(a_i)$. Next for each $i \in \mathcal{I}(\varphi)$ we define $Q_i = \{g \in G_o(t,\mathcal{W}) : g \cap apex(a_i) \neq \emptyset, g \cap apex(c_i) \neq \emptyset\}$. For each $i \in \cup \mathcal{I}(\mathcal{R})$ and each $q_i \in Q_i$ we construct a PRNS of $q_i \cap apex(a_i)$, say $\mathcal{P}_{q_i a_i}$, and a PRNS of $q_i \cap apex(c_i)$, say $\mathcal{P}_{q_i c_i}$. Next we define the set of conditional demands $\mathcal{C}_{oa}(\mathcal{W})$ = "for each $i \in \cup \mathcal{I}(\mathcal{R})$ and each $q_i \in Q_i$ $\mathcal{P}_{q_i a_i}$ is applied only for $q_i \cap apex(a_i)$ and the application order is: first $\mathcal{P}_{q_i a_i}$ then $\mathcal{W}$ ". We define PRNS

$\mathcal{W}_{oa} = \cup (\mathcal{P}_{q_i a_i} \cup \mathcal{W} : i \in \cup \mathcal{I}(\mathcal{R}), q_i \in Q_i, \mathcal{C}_{oa}(\mathcal{W}))$.

Now let $d_{q_i}$ be such a representative of such a net class that $(q_i \cap apex(a_i))\mathcal{P}_{q_i a_i}\hat{\ }$ is the context of the $d_{q_i}$ for $(q_i \cap apex(c_i))\mathcal{P}_{q_i c_i}\hat{\ }$. Let $\mathcal{P}_{b_i}$ be a PRNS of $b_i$, $i \in \cup \mathcal{I}(\mathcal{R})$, and

$\mathcal{C}_{ob}(\mathcal{W})$ = "for each $i \in \cup \mathcal{I}(\mathcal{R})$ and each $b_i \in B_i$ $\mathcal{P}_{b_i}$ is applied exclusively for $b_i$ in the position where rule preform $a_i \rightarrow b_i$ has transformed it and the application order is: first $\mathcal{P}_{b_i}$ then $\mathcal{W}$ "

be a set of conditional demands. Let us define PRNS

$\mathcal{W}_{oc} = \cup (\mathcal{P}_{q_i c_i} \cup \mathcal{W} : i \in \cup \mathcal{I}(\mathcal{R}), q_i \in Q_i, \mathcal{C}_{oc}(\mathcal{W}))$, where

$\mathcal{C}_{oc}(\mathcal{W})$ = "for each $i \in \cup \mathcal{I}(\mathcal{R})$ and each $q_i \in Q_i$ $\mathcal{P}_{q_i c_i}$ is applied only for $q \cap apex(c_i)$ and the application order is: first $\mathcal{P}_{q_i c_i}$ then $\mathcal{W}$ " is a set of conditional demands. Further we define PRNS $\mathcal{W}_o = \cup (\mathcal{P}_{b_i} \cup \mathcal{P}_{q_i c_i} \cup \mathcal{W} : i \in \cup \mathcal{I}(\mathcal{R}), \mathcal{C}_{ob}(\mathcal{W}), \mathcal{C}_{oc}(\mathcal{W}))$. Now we can give for the first rule preform application desired RNS $\mathscr{R}_{i\mathcal{W}_o} = \{q_i\mathcal{W}\hat{} \to d_{q_i}, \text{apex}(a_i)\mathcal{W}_{oa}\hat{} \to \{\text{apex}(b_i)\mathcal{P}_{b_i}\hat{}: b_i \in B_i\} : q_i \in Q_i\}, i \in \cup \mathcal{I}(\mathscr{R})$.

Now we obtain

$t\, \mathcal{W}\hat{}\, \mathscr{R}_{i\mathcal{W}_o}\hat{}\, (\mathcal{W}_o^{-1})\hat{} = t\, (a_i \to B_i)$, $i \in \cup \mathcal{I}(\mathscr{R})$, because $\mathcal{P}_{b_i}\hat{}$, $i \in \cup \mathcal{I}(\mathscr{R})$, are manoeuvre mightiness saving. In the next phase we continue the process for net $t\, \mathcal{W}\hat{}\, \mathscr{R}_{i\mathcal{W}_o}\hat{}\, (\mathcal{W}_o^{-1})\hat{}$ and obtain $t\, \mathcal{W}\hat{}\, \mathscr{R}_{i\mathcal{W}_o}\hat{}\, (\mathcal{W}_o^{-1})\hat{}\, \mathcal{W}_o\hat{}\, \mathscr{R}_{j\mathcal{W}_1}\hat{}\, (\mathcal{W}_1^{-1})\hat{} = t\, (a_i \to B_i)\, (a_j \to B_j)$, where $i,j \in \cup \mathcal{I}(\mathscr{R})$, and $\mathscr{R}_{j\mathcal{W}_1}$ and $\mathcal{W}_1$ are constructed for net $t\, \mathcal{W}\hat{}\, \mathscr{R}_{i\mathcal{W}_o}\hat{}\, (\mathcal{W}_o^{-1})\hat{}$ analogously with $\mathscr{R}_{i\mathcal{W}_o}$ and $\mathcal{W}_o$ for t. The continuation of that process concludes our proof for the implicit part of the theory.

2°    Now we are ready to move to prove the explicit interpretation of our equation. Let us denote $\varphi = \{\alpha_i \to \mathcal{B}_i : i \in \mathcal{I}(\varphi)\}$, whenever $\varphi \in \mathscr{R}_\mathcal{W}$. Now we have $\mathcal{W}$ and $\mathscr{R}_\mathcal{W}$ given. For each $i \in \cup \mathcal{I}(\mathscr{R}_\mathcal{W})$ let $\mathcal{P}_{\beta_i}$ be such a PRNS via which each $\beta_i \in \mathcal{B}_i$ is a concept. We construct a set of conditional demands $\mathcal{C}_{o\beta}(\mathcal{W}) =$ "for each $i \in \cup \mathcal{I}(\mathscr{R}_\mathcal{W})$ and each $\beta_i \in \mathcal{B}_i$ $\mathcal{P}_{\beta_i}$ is applied exclusively for $\beta_i$ in the position where rule preform $\alpha_i \to \beta_i$ has transformed it and the application order is: first $\mathcal{P}_{\beta_i}$ then $\mathcal{W}$. Further we define PRNS $\mathcal{W}_o = \cup (\mathcal{P}_{\beta_i} \cup \mathcal{W} : i \in \cup \mathcal{I}(\mathscr{R}_\mathcal{W}), \mathcal{C}_{o\beta}(\mathcal{W}))$, and give

$\mathscr{R}_i = \{\alpha_i(\mathcal{W}^{-1})\hat{} \to \{\beta_i(\mathcal{P}_{\beta_i}^{-1})\hat{}: \beta_i \in \mathcal{B}_i\}\}$. Now we can proceed as in 1°. □

It is worthy to observe that any macro/micro depend only on its micros/macros respectively and on the intervening PRNS´s, but not on the rewrite objects which might contain large number or even unlimited number of places for redexes of rules in micros.

## 3.3. Parallel Process and the Closure of Abstract Languages

**Definition 3.3.1.** Let $\mathcal{I}$ be an arbitrary set and for each $i,j \in \mathcal{I}$ let $\theta_{ij}$ be the abstraction relation, and let

$\tilde{\theta} = \Pi(\theta_{ij} : (i,j) \in \mathcal{I})$, thus $\tilde{\theta}$-associated languages is called $\mathcal{I}$-*abstract language*.

**Definition 3.3.2.** MACRO and MICRO TD. Let $\mathscr{R}$ be a set of RNS´s and $\mathscr{R}$ a TD over $\mathscr{R}$ (here singleton set of RNS´s and its element are equalized). We define a *macro TD* of $\mathscr{R}$ in regard to set $\mathscr{W}^o$ of interacting PRNS´s, denoted $\mathscr{R}_{\mathscr{W}^o}$, for which $\mathscr{R}_{\mathscr{W}^o} = \mathscr{R}(\mathscr{R} \leftarrow \mathscr{R}_{\mathscr{W}} : \mathscr{R} \in \mathscr{R}, \mathscr{W} \in \mathscr{W}^o)$, where $\mathscr{R}_{\mathscr{W}}$ is a macro RNS for $\mathscr{R}$ in regard to $\mathscr{W}$. We say that $\mathscr{R}$ is a *micro TD* of $\mathscr{R}_{\mathscr{W}^o}$, and denote it $(\mathscr{R}_{\mathscr{W}^o})_{\mathscr{W}^o{-1}}$.

The following "parallel"-theorem, one of the direct consequences from "altering macro RNS"-theorem, describes the invariability of the abstraction relation or the closures of abstract languages in class transformation relations, and taking advantage of the equation of "altering macro RNS"-theorem it gives TD-solutions for any problem each mother net of the subject of the problem is an abstract sister to a net which is a mother net of the subject of a problem TD-solutions of which are known. Cf. *class rewriting or confluence modulo* Jantzen M (1997), or TD with possibly freely chosen rules in RNS´s as action cf. *simulation* (Baeten JCM, Basten T (2001); van Glabbeek RJ (2001)), *bisimilarity* Aceto L, Fokkink WJ, Verhoef C (2001). It is worth to mention that there is close connections to game theories, inferring *winning game graphs* Thomas W (1997), *bisimulation equivalence* Burkart O, Caucal D, Moller F, Steffen B (2001), representation changes, abstraction and reformulation in artificial intelligence (Zucker J-D (2003); Holte RC, Choueiry BY (2003)).

**Theorem 3.3.** " Parallel theorem " . Let $\mathscr{R}$ be a RNS, $\theta$ the abstraction relation, a and b two such jungles that a$\theta$b, $\mathscr{W}_a$ and $\mathscr{W}_b$ two PRNS´s of such net c that a is a concept of c via $\mathscr{W}_a$ and b a concept of c via $\mathscr{W}_b$. Then we have a valid confluence condition regarding $\theta$ as follows:

   $1°$   a$\mathscr{R}\hat{}\,\,\theta\,\, b(\mathscr{R}_{\mathscr{W}_a{-1}})_{\mathscr{W}_b}\hat{}$ ,

and

   $2°$   a$\mathscr{R}_{\mathscr{W}_a}\hat{}\,\,\theta\,\, b\mathscr{R}_{\mathscr{W}_b}\hat{}$.

We call $\mathscr{R}$ and $(\mathscr{R}_{\mathscr{W}_a{-1}})_{\mathscr{W}_b}$ *parallel* with each other, and on the other hand consequently $\mathscr{R}_{\mathscr{W}_a}$ and $\mathscr{R}_{\mathscr{W}_b}$ are also parallel with each other, pairwise preserving $\theta$-classes in derivations.

# 3.4. Abstract Algebras

**Lemma 3.4.1.** All nets in any denumerable class of the abstraction relation have the shared substance (the *centre* of that class).

PROOF. Let $\theta$ be the abstraction relation and let H be a denumerable $\theta$-class. Each substance and its concepts are in the same $\theta$-class in according to corollary 3.1.1. Because H is an equivalence class being due to corollary 3.1.2, all substances in H are in $\theta$-relation with each other. Repeating the reasoning above for substances of substances and presuming that H is denumerable we will finally obtain the claim of the lemma. $\square$

**Lemma 3.4.2.** Let $\theta$ be the abstraction relation. Furthermore let $\mathscr{R}$ be a RNS, and let Q be a distinctive denumerable $\theta$-class with c being its centre. In addition we define a set of macro TD´s:

$$\mathfrak{R} = \{ \mathscr{R}_{\mathcal{W}}{}^*: \mathcal{W} \text{ is a PRNS of c } \}.$$

Therefore

$$\cup(Q\,\mathfrak{R}\,\theta) = c(\mathscr{R}\,\hat{}\cup\mathcal{I})\theta, \text{ where } \mathcal{I} \text{ is a trivial TD}.$$

PROOF. Because Q is distinctive, for each PRNS $\mathcal{W}$ of c $\mathscr{R}_{\mathcal{W}}$ has redexes exactly in one net of Q and the other nets in Q are in IRR($\mathscr{R}_{\mathcal{W}}$), our Parallel theorem yields $Q\mathscr{R}_{\mathcal{W}}{}^* \subseteq c(\mathscr{R}\,\hat{}\cup\mathcal{I})\theta$. Because $\theta$ is an equivalence relation, we get $Q\mathscr{R}_{\mathcal{W}}{}^*\theta = c(\mathscr{R}\,\hat{}\cup\mathcal{I})\theta$ and further $Q\mathfrak{R} = c(\mathscr{R}\,\hat{}\cup\mathcal{I})\theta$. $\square$

**Theorem 3.4.** "Abstraction closure-theorem".
Let A be the set of the denumerable $\theta$-classes, $\mathcal{R}$ is a set of RNS´s and

$$\mathscr{CR} = \cup(\,\{ \mathscr{R}_{\mathcal{W}}{}^*: \mathcal{W} \text{ is a PRNS of c } \} : \mathscr{R} \in \mathcal{R}, \text{ c is the centre of Q, } Q \in A\,)$$

be a union of macro TD´s liable to A-classes. Then if $\theta$ is the distinctive abstraction relation, pair (A, $\mathscr{CR}$) is an algebra, named *abstract algebra* or *net class rewriting algebra*.

PROOF.  Lemma 3.4.2 yields our claim, because our presumption for the abstraction relation yields each macro TD set { $\mathscr{R}_{\mathcal{W}}*$ : $\mathcal{W}$ is a PRNS of c } in $\mathscr{R}$ is matching exactly one θ-class, and because the construction of macro yields for each center c of A equation

|c$\mathcal{W}$ˆ $\mathscr{R}_{\mathcal{W}}$ˆ| = |c $\mathscr{R}$ˆ| and therefore consequently for each (Q∈A)  Q$\mathscr{R}$ is denumerable.  □

**Corollary 3.4.**  Parallel and Abstraction closure theorems are valid also in cases where the micro RNS is actually a general TD over a set of them and the intervening PRNS is a TD over a set of them as set in definition 3.3.2 (micro - macro TD).

In the next chapter we generalize the idea of PRNS to CRNS, "the cover RNS" where left sides of rule preforms are allowed to match apexes of right sides, and we study how nets are changed under TD´s over CRNS´s, and afterwards turned to be expressed by TD´s over PRNS´s of those rewrite objects. CRNS´s are important in expanding processes to search existing solutions in memory, the subjects of which being in the abstraction relation with the subject of the problem given to be solved.

4.§   <u>Type wise Problem Solving Regarding to Intervening RNS´s</u>

4.1.                  Cover RNS

In the following we are searching solutions for problems the mother nets having been built up by certain type of parts (elements in covers), this requirement is embedded in cover RNS´s, devoting CRNS as an abbreviation for that particular type of RNS. The apexes of the left sides of the rules in RNS´s in known TD (e.g. the catenation closure of RNS´s) may not be elements in any partition of the mother net of the problem studied, but in some more general cover. Furthermore we expand studies of RNS´s possessing multidimensional rules (G-RNS). The relations between PRNS and GCRNS are especially in focus. We construct generalized macro/micro (GMA/GMI) TD for GCRNS. Abstraction relation θ is then defined as before except PRNS is replaced with different variations of GCRNS.

**Definition 4.1.1.** For each relation λ we define *relation RNS of* λ, RNS(λ), such that

$$RNS(\lambda) = \{s \to T : s \in Dom(\lambda), T = s\lambda\}.$$

Notice that in general there is in force equation $(RNS(\lambda))^{-1} = RNS(\lambda^{-1})$.

**Definition 4.1.2.** COVER RNS. RNS $\mathscr{R}$ is a *cover RNS* (CRNS) of jungle s, if it fulfils conditions (i)-(v):

(i)  $\mathscr{R}$ is manoeuvre mightiness and arity mightiness saving, not instance sensitive,

(ii)  $L(apex(\bigcup(right(\mathscr{R}))))\backslash\Xi$ and set $L(s)$ are distinct with each other,

(iii) There is such jungle s´ for which $s \subseteq enc(s´)$ and

$\mathcal{C}(\mathscr{R}) \supseteq \{L(s´) \cap L(s´\mathscr{R}\hat{\,}) = \emptyset\}$ (*totally changing the ranked letters of* s) and

each rule preform of $\mathscr{R}$ has a redex in s´,

(iv) $\{(left(r), right(r)) : r \in \omega, \omega \in \mathscr{R}\}$ is an injection,

(v) the right side of each rule preform of $\mathscr{R}$ is a singleton.

Be notified that s itself may not possess any redex for CRNS. The set of all CRNS´s of jungle s is denoted CRNS(s). Observe that PRNS´s are examples of CRNS´s. We say that $s\mathscr{R}\hat{\,}$ is $\mathscr{R}$-*cover result* for s.

**Proposition 4.1.1.** "Characterization Clause". Let a and b be two distinct jungles. Then

$$\delta_D(a) = \delta_D(b) \Leftrightarrow \text{ there is such CRNS } \mathscr{R} \text{ that } a\mathscr{R}\hat{\,} = b.$$

PROOF. ´⇐´: CRNS is arity mightiness and manoeuvre mightiness saving, and therefore in the rewrite objects for CRNS the cardinality of the set of the outward linkage connections of the redexes is not changing in derivations.

PROOF. ´⇒´: Choose $\mathscr{R} = \{a \to b\}$. □

Next we concentrate to make notions adequate to differentiate PRNS and CRNS.

Clearly CRNS is a genuine generalization of PRNS, because PRNS´s do not allow ranked letter mightiness increase and redexes are limited to inside of rewrite objects and genuine overlapping between left and right sides of the rules are excluded.

Because a CRNS rule may have more than one ranked letter in the right side with e.g. different number of inside links within the right side than in the left side, then the family of the unoccupied arity sets of the ranked letters in a rewrite result may deviate from the family of the unoccupied arity set of any partition of the corresponding rewrite object and therefore a CRNS result may not be derived from the same rewrite object by any PRNS.

Proposition 3.1 and the greater expansive nature of CRNS compared to PRNS raise the question: For which jungle a and CRNS $\mathscr{R}$ of it there is such PRNS $\mathscr{W}$ of a that $a\mathscr{R}\hat{} = a\mathscr{W}\hat{}$? The next proposition gives an answer.

**Proposition. 4.1.2.** Let t be an arbitrary jungle. Let $\mathscr{R}$ be a *left-right distinct* CRNS of t (that is: for each rule preform $r$ apex(left($r$)) and apex(right($r$)) are distinct from each other), and for each rule preform $r$ in $\mathscr{R}$ let

$(\exists\,P\in\text{Par}(\text{apex}(\text{left}(r))))\,(\forall n\in\{\delta_D(\alpha) : \alpha\in\text{enc}(\text{apex}(\text{right}(r))), L(\alpha)\setminus\Xi \text{ is a singleton}\}\cup\{\delta_D(t) : t\in P\})$

$|\bigcup(p(P,t) : \delta_D(t) = n, t\in P)| = |\{c : |L(c)\cap\Xi| = n, c\in\text{enc}(\text{apex}(\text{right}(r))), L(\text{apex}(\text{right}(r)))\setminus\Xi \text{ is a singleton}\}|$.

Hence there is such PRNS $\mathscr{W}$ that $t\mathscr{R}\hat{} = t\mathscr{W}\hat{}$.

PROOF. We apply characterization proposition 3.1 upon the pairs of the left-right sides of the rule preforms in $\mathscr{R}$. Being due to our presumptions for the rules of $\mathscr{R}$ proposition 3.1 yields that for each rule preform $r$ in $\mathscr{R}$ there is such PRNS $\mathscr{W}_r$ that apex(right($r$)) is $\mathscr{W}_r$-partition result for apex(left($r$)), furthermore we require that all sets L(right($\mathscr{W}_r$)), $r\in\omega$, $\omega\in\mathscr{R}$, are distinct from each other. By choosing $\mathscr{W} = \bigcup(\mathscr{W}_r : r\in\omega, \omega\in\mathscr{R}, \mathcal{C} = \{\mathcal{C}(\mathscr{W}_r): r\in\omega, \omega\in\mathscr{R}\})$ we´ll get a desired PRNS, because $\mathscr{R}$ is left-right distinct (apex(left($\mathscr{R}$)) being a subset of a partition of t). □

In the following definition by generalization we give new types of RNS´s relating to the types of PRNS and CRNS.

**Definition 4.1.3.** GPRNS and GCRNS. GPRNS is RNS which is defined as PRNS but the condition "manoeuvre mightiness saving" is replaced with demand "not manoeuvre deleting and the right sides of the rule preforms are allowed to be also jungles instead of only nets", and GCRNS is RNS which is defined as CRNS with the above replacement.

Clearly we can generalize theorem 3.2 to be valid also for GPRNS in addition to PRNS.

**Proposition 4.1.3.** Let $\mathscr{R}$ be a GCRNS of jungle a. If the right sides of the rule preforms among the rules in $\mathscr{R}$ are distinct from each other (we say $\mathscr{R}$ is *distinct from right sides*) (reserving the symbols $C_d$RNS for CRNS and $GC_d$RNS for GCRNS in this respect), then

$$a\mathscr{R}^{\hat{}} \mathscr{R}^{-1\hat{}} = a.$$

If $\mathscr{R}$ is not distinct from right sides, then we have $a \subseteq a\mathscr{R}^{\hat{}} \mathscr{R}^{-1\hat{}}$.

PROOF. $GC_d$RNS is not manoeuvre deleting and is totally changing the ranked letters in rewrite objects (condition (iii) in the definition of CRNS). □

Next in the following chapter we prove "Altering Macro RNS"-theorem 3.2 generalized to deal also with the wider intervening RNS-type, cross colouring RNS, and in order to extend problem solving to fit also to that intervening type, a characterization of abstraction relation regarding that type is introduced.

## 4.2. Generalizing Altering Macro RNS Theorem

Before going to the next theorem we widen notion CRNS embedding it into general RNS´s at overlapping sections between left and right sides of rule preforms.

**Definition 4.2.1.** CROSS COLOURING RNS, CLCRNS. Let $\mathcal{W}$ be a RNS. For each net r we define relations $OL_r$ from pairs $(s,\mathcal{R})$ to nets, where s is a rule preform in $\mathcal{W}$ and $\mathcal{R}$ is a $GC_dRNS$ recursively:

$OL_r(s,\mathcal{R}_s) = \bigcup((apex(left(s))\cap apex(r))\mathcal{R}_s\hat{\ })$,

$OL_r(t,\mathcal{R}_t) = \bigcup((apex(left(t))\cap OL_r(s,\mathcal{R}_s))\mathcal{R}_t\hat{\ })$.

Notice that $OL_r$ may not be any mapping due to its potentiality to possess multi-images. We say that $\mathcal{W}$ is a *cross colouring RNS in respect to net* r, CLRNS(r), if $OL_r(t, \mathcal{R}_t)$ is an enclosement of apex(right(t)), whenever $apex(left(t))\cap OL_r(s,\mathcal{R}_s) \neq \varnothing$ for some s ($OL_r(t, \mathcal{R}_t)$ thus entitled a *coloured jungle* whereas $\mathcal{R}_t$ is a *colouring* $GC_dRNS$ of $\mathcal{W}$ in respect to r). If there is such a r-embedding net t that $PI(\bigcup(apex(left(\mathcal{W})))\cap t) \in Par(t)$, we say that CLRNS(r) $\mathcal{W}$ is *total*.

**Definition 4.2.2.** MACRO AND MICRO IN REGARD TO GPRNS AND CLRNS.

Let $\mathcal{R}$ be a RNS and $\mathcal{W}$ a nonconditional RNS of type T, T$\in$\{GPRNS,CLRNS\}. If there is such a RNS, $\mathcal{R}_\mathcal{W}$, and such a nonconditional T-type RNS $\mathcal{W}_o$ that there is in force an implicit equation of first order for unknown $\mathcal{R}_\mathcal{W}\hat{\ }$, thus $\mathcal{R}_\mathcal{W}$ being a composer for a linear composition of $\mathcal{R}\hat{\ }$:

$$\mathcal{W}\hat{\ }\,\mathcal{R}_\mathcal{W}\hat{\ }\,(\mathcal{W}_o^{-1})\hat{\ } = \mathcal{R}\hat{\ }.$$

we call $\mathcal{R}_\mathcal{W}$ *a macro* of $\mathcal{R}$ in regard to $\mathcal{W}$, indicated by MA($\mathcal{R},\mathcal{W}$). Consequently we entitle $\mathcal{R}$ *a micro* of $\mathcal{R}_\mathcal{W}$ in regard to $\mathcal{W}$, indicated by MI($\mathcal{R},\mathcal{W}$).

**Theorem 4.2.1.** Let r be a net and $\mathcal{R}$ be a RNS. Furthermore let $\mathcal{W}$ be a nonconditional total CLRNS(r) and

$a_{\mathcal{W}r} = \cap(t : r\in enc(t), PI(\bigcup(apex(left(\mathcal{W})))\cap t) \in Par(t))$,

" the smallest r-embedding net possessing a partition of r by $\mathcal{W}$ ", then such a partition of $a_{\mathcal{W}r}\mathcal{W}\hat{\,}$, say P, is achieved via $OL_r$-relations in $\mathcal{W}$ that $P\mathcal{W}^{-1}\hat{\,}$ is a partition of r and we can obtain MI($\mathcal{R}$,$\mathcal{W}$) as well as MA. Notice that in addition in a special case where each colouring $GC_dRNS$ in respect to net r in $\mathcal{W}$ can be chosen among the set of GPRNS´s, and if $\mathcal{P}$ is the set of those GPRNS´s, then $\cup\mathcal{P}$ is a GPRNS of r.

PROOF.   Analogous with altering macro theorem, the set of the colouring $GC_dRNS$´s equating GPRNS´s for elements of $a_{\mathcal{W}r}$-partitions as rewrite objects. □

**Definition 4.2.3.** MACRO AND MICRO IN REGARD TO TD OVER GPRNS´s AND CLRNS´s.
We define *macro* MA and *micro* MI in regard to TD over GPRNS´s, respectively over CLRNS´s as previously in the cases over PRNS´s. Consequently we use notations MA(TD,CLRNS) and respectively for MI. Furthermore for each TD $\mathcal{R}$ we denote

$\mathcal{C}_{\mathcal{R}+}(T) = \{\mathcal{R}_{\mathcal{W}^o} : \text{the elements of } \mathcal{W}^o \text{ are of type T}\}$ and

$\mathcal{C}_{\mathcal{R}-}(T) = \{(\mathcal{R}_{\mathcal{W}^o})_{\mathcal{W}^{o-1}} : \text{the elements of } \mathcal{W}^o \text{ are of type T }\}$,

whenever $T \in \{GPRNS, CLRNS\}$. TD´s $\mathcal{R}$ and $(\mathcal{R}_{\mathcal{W}^o})_{\mathcal{W}^{o-1}}$ are called *parallel* with each other, denoted also parallel($\mathcal{R}$) = $(\mathcal{R}_{\mathcal{W}^o})_{\mathcal{W}^{o-1}}$ or parallel($(\mathcal{R}_{\mathcal{W}^o})_{\mathcal{W}^{o-1}}$) = $\mathcal{R}$.

Notice that because CLRNS´s are genuine generalizations of GPRNS´s we have equations

$\mathcal{C}_{\mathcal{R}+}(GPRNS) \subset \mathcal{C}_{\mathcal{R}+}(CLRNS)$   and   $\mathcal{C}_{\mathcal{R}-}(GPRNS) \subset \mathcal{C}_{\mathcal{R}-}(CLRNS)$ .

Theorem 4.2.1 yields the following theorem for more general cases:

**Theorem 4.2.2.**  For each nonconditional $\mathcal{W}$ of type T, $T \in \{GPRNS, CLRNS\}$, and each string $\mathcal{R}$ over set of RNS´s, there is $\mathcal{R}_{\mathcal{W}}$, and such of type T RNS $\mathcal{W}_o$ that

$$\mathcal{W}\hat{\,} \; \mathcal{R}_{\mathcal{W}}\hat{\,} \; (\mathcal{W}_o^{-1})\hat{\,} \; = \; \mathcal{R}\hat{\,} \; .$$

**Corollary 4.2.1.** The above theorem can be dressed also somewhat more generally:

Let $\mathcal{R}$ be a micro TD over set of RNS´s and $\mathcal{W}^o$ a set of intervening nonconditional RNS´s of type T. Then there is such macro TD of $\mathcal{R}$, $\mathcal{R}_M$, and such set of reversed T-type RNS´s, $\mathcal{W}^o_o$, that we have commutativity condition

$$\mathcal{W}^o \hat{\ } \mathcal{R}_M \hat{\ } \mathcal{W}^o_o \hat{\ } = \mathcal{R} \hat{\ },$$

and it manifestates a natural transformation between Functors determining parallel rewriting cf. Theorem 3.3.

**Definition 4.2.4.** GENERALIZED ABSTRACTION RELATION. The *generalized abstraction relation* in regard to type T of intervening RNS, denoted GAR(T), T∈{PRNS,GPRNS,CLRNS,GC$_d$RNS,GCRNS}, (in short *abstraction relation of type* T ) is such a binary relation in the set of the nets, where for each pair (here (s,t)) there is such net c and intervening RNS $\mathcal{W}_1$ and $\mathcal{W}_2$ of type T, that

$$c\mathcal{W}_1 \hat{\ } = s \quad \text{and} \quad c\mathcal{W}_2 \hat{\ } = t .$$

Nets s and t are said to be *abstract sisters of type* T with each other, c being a common substance of s and t. Notice that GAR is a genuine generalization for abstraction relation AR, and that AR = GAR(PRNS).

**Proposition 4.2.1.** "A characterization of abstraction relation GAR(CLRNS)".

Let a and b be two nets and let θ be GAR(CLRNS). Then

$$a \theta b \Leftrightarrow \delta_D(a) = \delta_D(b) .$$

PROOF.  Theorem 4.2.1 and characterization proposition 4.1.1. □

**Remark 4.2.** Straightforwardly widening the definition for "parallel" to deal with intervening RNS´s of type GPRNS and CLRNS instead of solely dealing with type PRNS, we clearly have the results for GAR(CLRNS) as is obtained for AR in corollaries 3.1.1 and 3.1.2, result 3.1, parallel-theorem, lemmas 3.4.1 and 3.4.2 and theorem 3.4 and finally consequently results concerning generalizations for TD´s as stated in Corollary 3.4.

**Proposition 4.2.2.** "Characterization of GAR".

Let T∈{PRNS,GPRNS,CLRNS,GC$_d$RNS,GCRNS} and let s and t be nets. Then s and t are abstract sisters of type T, if and only if there exist such intervening RNS $\mathcal{V}_s$ and $\mathcal{V}_t$ of type T that ($\exists$ $A_s \in Par(s(\mho_s^{-1})\hat{\,})$) and ($\exists$ $A_t \in Par(t(\mho_t^{-1})\hat{\,})$) there is a bijection between $A_s$ and $A_t$.

PROOF. The right sides of rule preforms must pair wise in both of the intervening RNS´s possess the same number of different manoeuvre letters liable to cardinality of the bijection between the related partitions. $\square$

**Definition 4.2.5.** CONGRUENCE. Let $\theta$ be a relation in the set of the jungles. We say that $\theta$ is a *congruent relation of TD- type* T, if there is in force:

a $\theta$ b $\Leftrightarrow$ a$\varphi_a$ $\theta$ b$\varphi_b$  whenever $\varphi_a$ and $\varphi_b$ are TD´s of type T.

Each congruent relation of type T, which is an equivalence relation, is entitled *congruence relation of type* T. The set of all congruence relations of type T is denoted Cg(T).

**Theorem 4.2.3.** For each TD-type $T_m$ GAR($T_m$)$\in$Cg($T_n$), m $\geq$ n, m, n = 1,2,3, where $(T_1, T_2, T_3)$ = (PRNS,GPRNS,CLRNS).

PROOF. GAR(T) is congruent, because any catenation of TD´s is of the same type as the TD of the most general type in that catenation and $T_m$ is a generalization of $T_n$, if m $\geq$ n. Proposition 4.2.1 yields the equivalence requirement. $\square$

SYNTAX OF AUTOMATED PROBLEM SOLVING SYSTEM.

The mother net of a given problem is first transformed by an intervening CLRNS to concept net for which we construct an abstract sister, one of the substances of which has a partition in a bijection with a partition of a substance of the said concept net. Now the known transducer renders possibility to construct a macro for it, the parallel counterpart and finally a micro parallel macro, because the reached concepts guarantee the survival of information of the rules in known TD´s in the process. By iteration we can reach for our original problem a presolution, which finally is a desired solution, if the product is in the anticipated language fulfilling the set of limit demands.

Directly searching a common substance of certain type for a net pair would be substantially more difficult if even impossible than going through pair (macro,parallel macro) in a case where either of the nets in said pair is undenumerable regarding to the cardinalities of the sets of their letters (and actually even if the cardinality of one of them is immense although denumerable).

## Conclusions

The present study represents a new way to describe knowledge with generalized universal algebra allowing loop structures so very important in AI languages and which gives an extensive variety of notional relations between net entities without restricting the semantic use. Consequently a new syntax model for solving problems defined by said nets is established flexibly utilizing notional similarities with original problems to further match solutions in memory data banks additionally creating transducer graphs of solving rewrite systems and thereof closure system of solving classes.

## For the future considerations

Conceptual graphs constitute equivalence classes as the form of elements in a closed quotient systems, meaning that parallel transformation applied to those classes inevitably drops images back into the set of those particular classes, which guarantees automated problem solving and consequently is in the interest of this research. For the reason of "memory hunting" it might be worthwhile to consider continuing the process of abstract net pair forming in the chain formation by intervening rewriting then asking if this kind of "catenation strings" form elements in some closed system. Furthermore type wise use of normal forms in renetting (especially creating new links by right side substitutions) raises a promising question of the types of the quotient closure itself.

## Acknowledgements

I own the unparalleled gratitude to my family, my wife and five children for the cordial environment so very essential on creative working.

**References**

Aceto L, Fokkink WJ, Verhoef C (2001) Structural Operational Semantics. In: Bergstra JA, Ponse A, Smolka SA, editors. Handbook of Process Algebra. Amsterdam: Elsevier. pp.197-292.

Baeten JCM, Basten T (2001) Partial-Order Process Algebra (and Its Relation to Petri Nets). In: Bergstra JA, Ponse A, Smolka SA, editors. Handbook of Process Algebra. Amsterdam: Elsevier. pp. 769-872.

Baeten JCM, Middelburg CA (2001) Process Algebra with Timing: Real Time and Discrete Time. In: Bergstra JA, Ponse A, Smolka SA, editors. Handbook of Process Algebra. Amsterdam: Elsevier. pp. 627-684.

Best E, Devillers R, Koutny M (2001) A Unified Model for Nets and Process Algebra. In: Bergstra JA, Ponse A, Smolka SA, editors. Handbook of Process Algebra. Amsterdam: Elsevier. pp. 873-944.

Burkart O, Caucal D, Moller F, Steffen B (2001) Verification on Infinite Structures. In: Bergstra JA, Ponse A, Smolka SA, editors. Handbook of Process Algebra. Amsterdam: Elsevier. pp. 545-623.

Burris S, Sankappanavar HP (1981) A Course in Universal Algebra. New York: Springer-Verlag. 276 p.

Chang CC, Keisler HJ (1973) Model theory. Amsterdam: North-Holland Publishing Company. 554 p.

Cleaveland R, Lüttgen G, Natarajan V (2001) Priority in Process Algebra. In: Bergstra JA, Ponse A, Smolka SA, editors. Handbook of Process Algebra. Amsterdam: Elsevier. pp. 711-765.

Denecke K, Wismat SL (2002) Universal algebra and the applications in theoretical Computer Science. Boca Raton: Chapman & Hall. 383 p.

Diekert V, Métivier Y (1997) Partial Commutation and Traces. Rozenberg G, Salomaa A, editors. In: Handbook of Formal Languages, Vol.3 Beyond Words. Berlin: Springer-Verlag. pp. 457-533.

Engelfriet J (1997) Context-Free Graph Grammars. In: Rozenberg G, Salomaa A, editors. Handbook of Formal Languages, Vol.3 Beyond Words. Berlin: Springer-Verlag. pp.125-213.

Gabbay DM, Hogger CJ, Robinson JA (1995) Handbook of Logic in Artificial Intelligence and Logic Programming, Vol 4 Epistemic and Temporal Reasoning. New York: Oxford University Press Inc. pp. 242-350.

Gécseg F, Steinby M (1997) Tree Languages. In: Rozenberg G, Salomaa A, editors. Handbook of Formal Languages, Vol.3 Beyond Words. Berlin: Springer-Verlag. pp.1-68.

Holte RC, Choueiry BY (2003) Abstraction and reformulation in artificial intelligence. Philos Trans R Soc Lond B Biol Sci July 29; 358(1435): 1197–1204.

Jantzen M (1997) Basic of Term Rewriting. In: Rozenberg G, Salomaa A, editors. Handbook of Formal Languages, Vol.3 Beyond Words. Berlin: Springer-Verlag. pp. 269-337.

Jonsson B, Yi W, Larsen KG (2001) Probabilistic Extensions of Process Algebra. In: Bergstra JA, Ponse A, Smolka SA, editors. Handbook of Process Algebra. Amsterdam: Elsevier. pp. 685-710.

Körner E, Gewalting M-O, Körner U, Richter A, Rodemann T (1999) Organisation of Computation in brain-like Systems, Neural Networks (special issue) 12 pp. 989-1005, New York: Elsevier Science.

Meseguer J, Goguen JA (1985) Initiality, Induction, and Computability. In: Nivat M, Reynolds JC, editors. Algebraic methods in semantics. London: Cambridge University Press. pp. 459-541.

Müller J (1997) A non-categorical Characterization of Sequential Independence for Algebraic Graph Rewriting and some Applications, Technische Universität Berlin, Forschunsberichte des Fachbereichs Informatik, Nr 97-18.

Nivat M, Reynolds JC, editors. (1985) Algebraic methods in semantics. London: Cambridge University Press. 634 p.

Ohlebusch E (2002) Advanced Topics in Term Rewriting. New York: Springer-Verlag. pp. 179-242, 327-338.

Rozenberg G, Salomaa A, editors. (1997) Handbook of Formal Languages, Vol.3 Beyond Words. Berlin: Springer-Verlag. 623 p.

Thomas W (1997) Language, Automata and Logic. In: Rozenberg G, Salomaa A, editors. Handbook of Formal Languages, Vol.3 Beyond Words. Berlin: Springer-Verlag. pp. 389-455.

Tirri S, Aurela AM (1989) Solution of nonlinear differential equations with picard iterants of fixed forms. Int J Comput Math, Vol. 27 Issue 1 pp. 33 – 54.

Tirri S (1990) The congruence theory of closure properties of regular tree languages. Theor Comput Sci, Vol 76 Issues 2-3, Nov 21, pp. 261-271.

Tirri SI (2009) Cover type controlled graph rewriting based parallel system for automated problem solving. US Patent Application, non-pending, 20090171876, Jul 2009.

van Glabbeek RJ (2001) The Linear Time-Branching Time Spectrum I. The Semantics of Concrete, Sequential Processes. In: Bergstra JA, Ponse A, Smolka SA, editors. Handbook of Process Algebra. Amsterdam: Elsevier. pp.3-100.

Zucker J-D (2003) A grounded theory of abstraction in artificial intelligence. Philos Trans R Soc Lond B Biol Sci July 29; 358(1435): 1293–1309.